\documentclass[10pt,twocolumn,letterpaper]{article}

\usepackage{iccv}              

\definecolor{iccvblue}{rgb}{0.21,0.49,0.74}
\usepackage[pagebackref,breaklinks,colorlinks,allcolors=iccvblue]{hyperref}

\usepackage[accsupp]{axessibility}
\usepackage{algorithm}
\usepackage{algcompatible}
\usepackage{algorithmicx}
\usepackage{algpseudocode}
\usepackage{multirow}

\def\paperID{12847} 
\def\confName{ICCV}
\def\confYear{2025}

\title{Diffusion-based 3D Hand Motion Recovery with Intuitive Physics}

\author{Yufei Zhang$^1$, Zijun Cui$^{2}$, Jeffrey O. Kephart$^3$, Qiang Ji$^1$ \\
$^1$Rensselaer Polytechnic Institute, $^2$Michigan State University, $^3$IBM Research
 \\
{\tt\small yufeizhang96@outlook.com, cuizijun@msu.edu, kephart@us.ibm.com,
jiq@rpi.edu}
}

\begin{document}
\maketitle
\begin{abstract}
While 3D hand reconstruction from monocular images has made significant progress, generating accurate and temporally coherent motion estimates from videos remains challenging, particularly during hand-object interactions. In this paper, we present a novel 3D hand motion recovery framework that enhances image-based reconstructions through a diffusion-based and physics-augmented motion refinement model. Our model captures the distribution of refined motion estimates conditioned on initial ones, generating improved sequences through an iterative denoising process. Instead of relying on scarce annotated video data, we train our model only using motion capture data without images. We identify valuable intuitive physics knowledge during hand-object interactions, including key motion states and their associated motion constraints. We effectively integrate these physical insights into our diffusion model to improve its performance. Extensive experiments demonstrate that our approach significantly improves various frame-wise reconstruction methods, achieving state-of-the-art (SOTA) performance on existing benchmarks.
\end{abstract}    
\section{Introduction}
\label{sec:intro}

\begin{figure}[t]
\begin{center}
\includegraphics[width=0.95\linewidth]{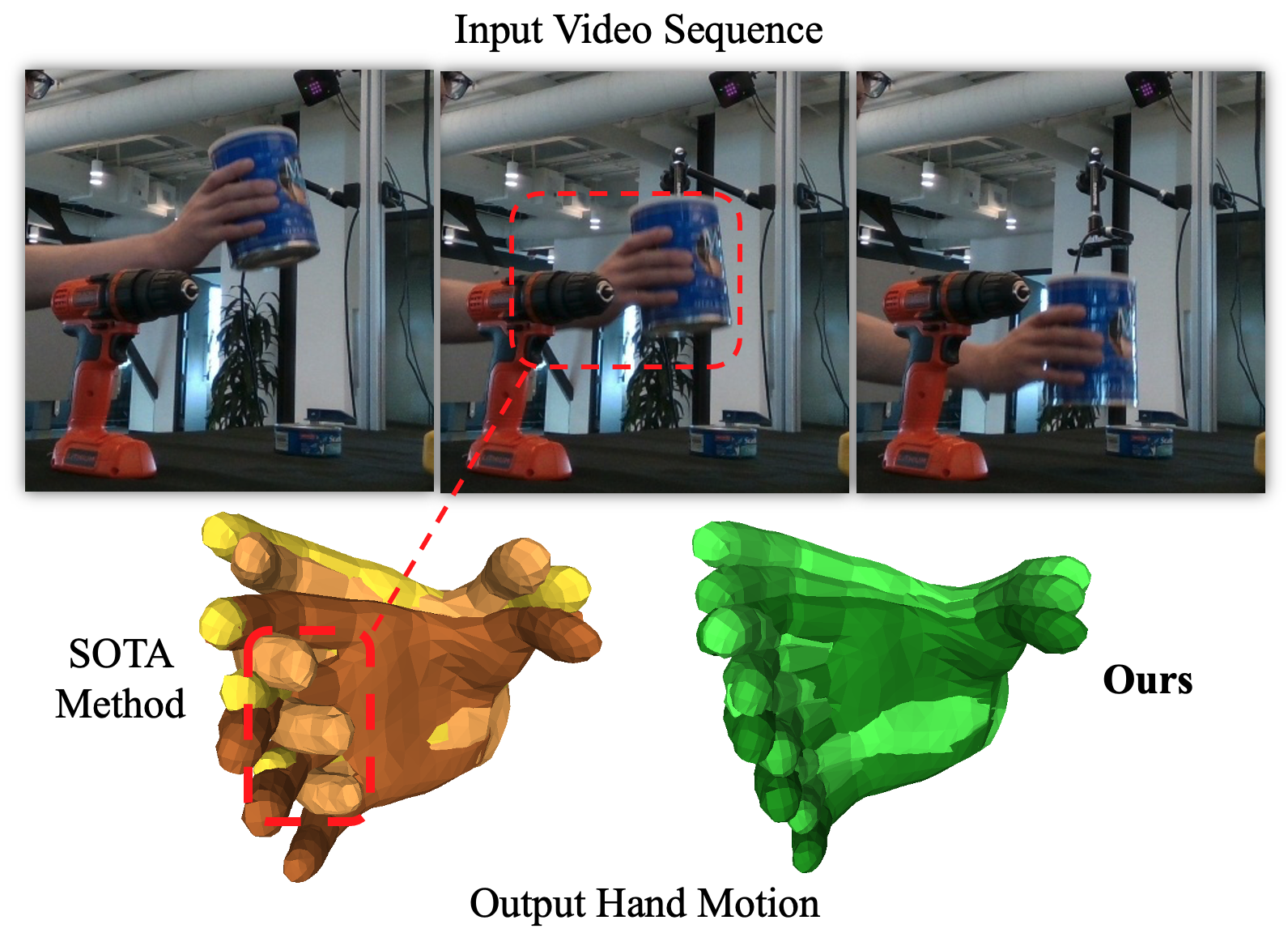}
\vspace{-0.3cm}
\caption{\textbf{Motivation for Temporal Dependency Modeling.} The example testing sequence (top) is from DexYCB~\cite{chao2021dexycb}. The 3D hand reconstructions are generated by the leading frame-wise reconstruction method HaMer~\cite{pavlakos2023reconstructing} (bottom left) and our method (bottom right). The 3D hand reconstructions are color-coded from light to dark across consecutive frames (left to right).}
\label{fig:ch6-intro}
\end{center}
\end{figure}

Reconstructing 3D hand and its motion from a single RGB camera has broad applications spanning Virtual/Augmented Reality (VR/AR) \cite{grubert2018effects,bai2020user}, robotic dexterous manipulation \cite{qin2022one}, and beyond. While deep learning has led to promising single-image hand reconstruction~\cite{zimmermann2017learning,yang2019aligning,hasson2020leveraging,lin2021end,spurr2021self,liu2021semi}, generating temporally coherent hand motion from video sequences remains challenging. The challenge is particularly pronounced during hand-object interactions, where severe occlusions may occur. As illustrated in Figure~\ref{fig:ch6-intro}, even leading frame-wise 3D hand reconstruction methods produce degraded predictions when hands are partially hidden by manipulated or external objects (marked in red).

To enhance model robustness and produce coherent motion recovery, video-based methods that leverage temporal information offer a promising direction. Traditional approaches require training on annotated video sequences, which are significantly scarcer and more labor-intensive to collect compared to single-frame annotations. This data scarcity becomes particularly problematic in hand-object interaction scenarios, where the variety of possible object manipulations further increases the annotation effort required. Consequently, despite their advantages in maintaining temporal coherence, existing video-based approaches have not yet achieved the per-frame reconstruction accuracy of state-of-the-art image-based methods~\cite{Fu_2023_ICCV,lee2023fourierhandflow,dong2024hamba}.

To tackle the limited availability of video data, recent works~\cite{baradel2022posebert,zhang2024physpt,zhou2022toch,xie2024ms,luo2024physics} propose training motion refinement models using only 3D motion data. These models refine initial 3D hand reconstructions $\mathbf{y}_{1:T}$ from image-based methods into temporally coherent sequences $\mathbf{x}_{1:T}$. However, this strategy faces two fundamental limitations: (1) their deterministic nature prevents them from capturing the inherent uncertainty in image-based estimates, thereby limiting their potential to generate more accurate reconstructions, and (2) they predominantly focus on specific types of hand-object interactions, neglecting the general physical principles that govern efficient hand-environment interactions.

In this work, we address these limitations by introducing a physics-augmented conditional diffusion model for hand motion refinement that jointly (1) captures uncertainty in initial frame-wise estimates through diffusion models and (2) incorporates principles of intuitive physics to generate natural hand dynamics during object interactions. Specifically, instead of building a deterministic mapping from $\mathbf{y}_{1:T}$ to $\mathbf{x}_{1:T}$, we model their conditional probabilistic distribution $p(\mathbf{x}_{1:T} | \mathbf{y}_{1:T})$. This distribution can be complex due to the intricate hand motions and various noisy image-based estimates. Inspired by recent advancements in diffusion models~\cite{ho2020denoising,song2020denoising}, we develop a diffusion-based framework to capture this complex distribution and generate refined estimates by iteratively denoising the noisy motion estimates toward more physically plausible ones. Moreover, we draw insights from studies on intuitive physics that reveal how hands gracefully behave in the physical world during object interactions~\cite{santello2013neural,zhao2013robust,bhatt2017hand}. Based on these insights, we identify four fundamental states in hand-object interactions: \textit{Reaching}, \textit{Stable Grasping}, \textit{Manipulation}, and \textit{Releasing}, each involving distinct motion patterns and physical constraints that ensure natural hand-object interaction. We explicitly incorporate these state-specific constraints into our diffusion model to enhance its ability to produce physically plausible and natural hand motions. By combining diffusion modeling with intuitive physics principles, we train our model solely on motion capture data. Once trained, it can be directly integrated with any existing frame-wise reconstruction method to enhance 3D hand motion recovery.

To summarize, our main contributions are as follows:
\begin{itemize} 
\item We introduce a new hand motion recovery framework based on conditional diffusion, which captures initial motion estimation uncertainty and generates refined motion estimates through iterative denoising.
\item We identify key intuitive knowledge of hand dynamics and effectively incorporate this knowledge into our diffusion model to enhance its dynamic modeling.
\item Through extensive experiments, we validate the effectiveness of our approach and demonstrate SOTA performance compared to existing motion refinement and video-based reconstruction methods.
\end{itemize}

\section{Related Work}
\label{sec:relatedwork}

\noindent\textbf{Image-based 3D Hand Reconstruction.} Reconstructing 3D hands from images has rapidly advanced in recent years due to deep learning. Regarding fully-supervised approaches, various strategies have been proposed to improve reconstruction accuracy, including the use of prior 3D hand models~\cite{baek2019pushing,zhang2019end,park2022handoccnet} and model designs for capturing spatial dependencies~\cite{ge20193d,choi2020pose2mesh,moon2020i2l,lin2021end,chen2022mobrecon,pavlakos2023reconstructing,zhou2024simple,Qi_2024_CVPR,yu2023overcoming,jiang2023probabilistic,Zhou_2024_CVPR,potamias2024wilor}. Additionally, some methods address specific challenges, such as hand-object interactions~\cite{hasson2019learning,tse2022collaborative,yang2022artiboost,xu2023h2onet,Ye_2024_CVPR,xu2024handbooster} or reconstruction of interacting hands~\cite{zhang2021interacting,li2022interacting,wang2023memahand,ren2023decoupled,yu2023acr,zuo2023reconstructing}. To reduce the reliance on expensive 3D annotations, some approaches have further explored weakly-supervised learning~\cite{boukhayma20193d,chen2023mhentropy,mueller2018ganerated,kulon2020weakly,spurr2020weakly,chen2021model,ren2022end,tu2023consistent,gao2022cyclehand,baek2020weakly,zhang2024weakly}. Nonetheless, current image-based methods often produce unrealistic motion estimates, especially under poor image conditions. A promising direction to address these issues is reconstructing 3D hands from video sequences, as reviewed below.

\noindent\textbf{Video-based 3D Hand Reconstruction.} Given a video, some approaches focus on extracting temporally consistent feature through sophisticated temporal models~\cite{choi2021beyond,Fu_2023_ICCV}. They rely heavily on annotated video sequences for training, which are inherently limited in availability and diversity due to the dependence on specialized motion capture systems. Consequently, their per-frame reconstruction accuracy often falls short of SOTA image-based methods~\cite{zhang2024physpt}. To alleviate the data scarcity issue, alternative approaches propose training with in-the-wild videos by incorporating photometric consistency~\cite{hasson2020leveraging} or motion smoothness~\cite{yang2020seqhand,kocabas2020vibe,liu2021semi,ziani2022tempclr,duran2024hmp}, enabling coherent 3D reconstruction across consecutive frames without requiring 3D annotations. Unlike these feature extraction approaches, PoseBERT~\cite{baradel2022posebert} achieves enhanced performance by leveraging Transformer~\cite{vaswani2017attention} trained with random input perturbation and masking. Once trained, it can be directly used to refine image-based predictions. However, PoseBERT is primarily deterministic and data-driven. Our work advances this refinement paradigm by explicitly modeling uncertainty in the initial estimates while effectively incorporating intuitive physics knowledge to improve hand dynamics modeling.

In summary, our work distinguishes itself from existing methods through the following key contributions: First, we introduce a novel video-based hand motion recovery framework based on a conditional diffusion model. While diffusion has been applied to image-based 3D hand reconstruction~\cite{li2024hhmr} and modeling hand affordance for certain objects~\cite{liu2024geneoh}, we formulate a conditional diffusion model for hand motion refinement. Second, inspired by the recent success of leveraging intuitive physics for 3D human body reconstruction~\cite{vo2020spatiotemporal,tripathi20233d}, we exploit intuitive physics knowledge in hand motion and integrate this valuable knowledge into our diffusion model to enhance its ability to capture realistic hand dynamics. Unlike object-specific modeling~\cite{zhou2022toch,hao2024hand,luo2024physics,zhu2024grip} or muscle dynamics~\cite{xie2024ms} studied by existing models, our approach to intuitive physics learns human intuitive understanding of natural hand movements and their interactions with the physical world.

\begin{figure*}[t]
\begin{center}
\includegraphics[width=0.8\linewidth]{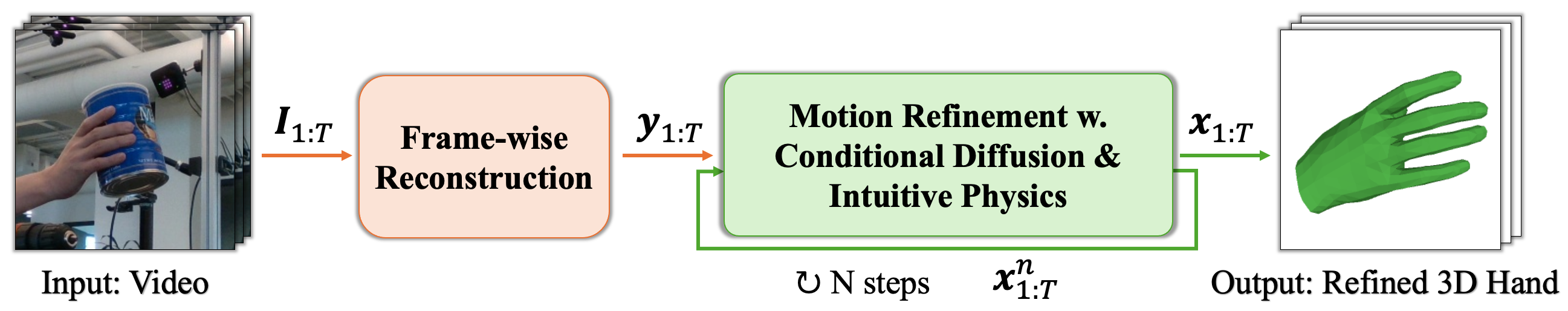}
\vspace{-0.1cm}
\caption{\textbf{Overview of the Proposed Method.} Given a sequence of hand image frames $\mathbf{I}_{1:T}$ with sequence length $T$, the frame-wise reconstruction model produces initial estimates of 3D hands for every frame $\mathbf{y}_{1:T}$. Then, the proposed motion refinement model, based on conditional diffusion and incorporated with intuitive physics, generates improved 3D hand motion estimates $\mathbf{x}_{1:T}$.}
\label{fig:overview}
\end{center}
\end{figure*}

\section{Method}

An overview of the proposed method for reconstructing 3D hand motion from a sequence of video frames is illustrated in Figure~\ref{fig:overview}. Our approach begins with a frame-wise reconstruction model, which generates per-frame 3D hand configurations, as detailed in Section~\ref{sec:framewise}. The frame-wise reconstruction model may struggle to capture temporal dependencies, leading to unrealistic motion estimates. We thus propose a novel motion refinement model that combines a conditional diffusion model and hand intuitive physics to generate more accurate motion estimates. The details of the diffusion-based refinement framework are provided in Section~\ref{sec:diffusionmodel}, followed by a discussion in Section~\ref{sec:intuitivephysics} on the role of intuitive physics in hand motion and how this knowledge is incorporated into the diffusion model.

\subsection{Frame-wise Reconstruction Model}
\label{sec:framewise}

The frame-wise reconstruction model takes as input $T$ video frames and outputs 3D hands for each frame. Our pipeline is agnostic to the specific choice of frame-wise reconstruction methods. While prior works have explored various frameworks to achieve state-of-the-art performance on different benchmarks, most rely on the widely used 3D hand representation, MANO~\cite{romero2017embodied}. Therefore, we consider frame-wise reconstruction models estimating MANO configurations.

MANO represents the hand surface using a 3D mesh $\mathbf{M}(\boldsymbol{\theta}, \boldsymbol{\beta}) \in \mathbb{R}^{778 \times 3}$, where $\boldsymbol{\theta} \in \mathbb{R}^{15 \times 3}$ and $\boldsymbol{\beta} \in \mathbb{R}^{10}$ denote the pose and shape parameters, respectively. Additionally, 3D hand joints $\mathbf{P}(\boldsymbol{\theta},\boldsymbol{\beta}) \in \mathbb{R}^{J\times3}$ can be computed as a linear combination of the mesh vertices through $\mathbf{P}(\boldsymbol{\theta},\boldsymbol{\beta}) = \mathbf{H}\mathbf{M}(\boldsymbol{\theta},\boldsymbol{\beta})$, where $\mathbf{H} \in \mathbb{R}^{J\times778}$ is a joint regressor learned from data, and $J = 21$ denotes the number of modeled hand joints. Using MANO as 3D hand representation, we employ a well-established method to estimate the hand pose and shape parameters, $\boldsymbol{\theta}_{1:T}$ and $\boldsymbol{\beta}_{1:T}$, for every frame in the input video sequence. Given these initial 3D hand configuration estimates, $\boldsymbol{y}_{1:T}=(\boldsymbol{\theta}_{1:T}, \boldsymbol{\beta}_{1:T})$, our proposed motion refinement model generates enhanced motion estimates, as detailed in the following section.

\subsection{Diffusion-based Motion Refinement}
\label{sec:diffusionmodel}

Given the initial 3D hand reconstructions, denoted as $\mathbf{y}_{1:T}$, existing approaches~\cite{baradel2022posebert,zhang2024physpt} adopt Transformer~\cite{vaswani2017attention} to directly learn a deterministic mapping from $\mathbf{y}_{1:T}$ to the refined motion estimates $\mathbf{x}_{1:T}$ using 3D hand motion capture data. Nonetheless, these initial estimates, $\boldsymbol{y}_{1:T}$, can present various issues such as motion jitter or degraded reconstruction due to occlusions. Instead of learning a deterministic mapping, we model the conditional distribution $p(\mathbf{x}_{1:T}|\mathbf{y}_{1:T})$. Particularly, motivated by the recent success of diffusion models~\cite{ho2020denoising,song2020denoising} in capturing complex distributions, we adopt a diffusion-based framework to represent the data distribution $q(\mathbf{x}_{1:T}|\mathbf{y}_{1:T})$. Specifically, like \cite{yue2024resshift,song2020denoising}, we formulate forward and inverse diffusion processes that connect the initial estimates $\boldsymbol{y}_{1:T}$ and the refined estimates $\boldsymbol{x}_{1:T}$, as detailed in the following sections.

\noindent\textbf{Forward Diffusion Process.} Let's denote $\mathbf{y}_{1:T}$ as the observed initial motion estimates and $\mathbf{x}_{1:T}$ as the final refined motion, which is clean hand motion data during training. The forward process progressively adds Gaussian noise to transition from $\mathbf{x}_{1:T}$ to $\mathbf{y}_{1:T}$ through $N$ diffusion steps. Specifically, denote the residual between the initial and refined motion estimates as $\mathbf{e}_{1:T}=\mathbf{y}_{1:T}-\mathbf{x}_{1:T}$. The marginal distribution at any forward diffusion step $n$ is formulated as:
\vspace{-0.32cm}
\begin{equation}
q(\mathbf{x}_{1:T}^n|\mathbf{x}_{1:T},\mathbf{y}_{1:T}) \sim \mathcal{N}(\mathbf{x}_{1:T}+\eta_n \mathbf{e}_{1:T}, \kappa^2 \eta_n \mathbf{I}),
\label{eq:forwarddiffusion}
\end{equation}
where $\kappa$ is a hyper-parameter controlling the noise variance, $\mathbf{I}$ is the identity matrix with the same size as $\mathbf{x}_{1:T}^n$, and $\{\eta_t\}_{t=1}^T$ is a predefined shifting sequence, which monotonically increases with the step $n$ and satisfies $\eta_1 \to 0$ and $\eta_N \to 1$. For the detailed derivation of the marginal distribution and the construction of the shifting sequence, please refer to \cite{yue2024resshift}. As illustrated in Equation~\ref{eq:forwarddiffusion}, the forward diffusion process starts from a Dirac delta distribution centered at $\mathbf{x}_{1:T}$ at step 0. The process gradually adds $\mathbf{e}_{1:T}$ and injects random noise, converging to a marginal distribution $\mathcal{N}(\mathbf{y}_{1:T},\kappa^2\mathbf{I})$ at diffusion step $N$. With this formulation, we establish a forward diffusion process that progressively transforms the true motion toward the initial motion estimates, using their residuals as the direction for noising while incorporating Gaussian noise. In the following, we describe the reverse diffusion process, which denoises the initial motion estimates back to the true motion.

\begin{figure}[t]
\begin{center}
\includegraphics[width=0.99\linewidth]{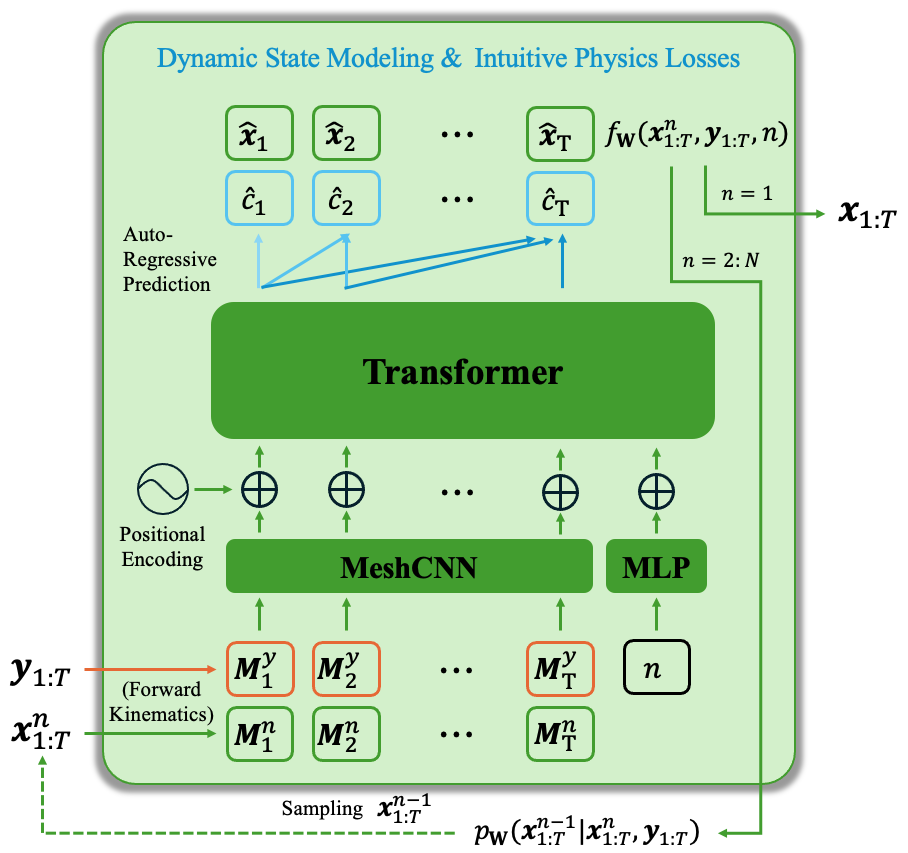}
\vspace{-0.2cm}
\caption{\textbf{Illustration of the Reverse Diffusion Framework.} The framework operates iteratively, where at step $n$, it processes the initial motion estimates $\mathbf{y}_{1:T}$ and the current sample $\mathbf{x}_{1:T}^n$ to generate the estimated clean motion $\hat{\mathbf{x}}_{1:T}=f_\mathbf{W}(\mathbf{x}_{1:T}^n,\mathbf{y}_{1:T},n)$. This estimation is achieved through a hybrid architecture combining MeshCNN and Transformer to capture spatial-temporal dependencies, along with an MLP for encoding diffusion step information. The estimated clean motion $\hat{\mathbf{x}}_{1:T}$ then parameterizes the transition distribution $p_\mathbf{W}(\mathbf{x}_{1:T}^{n-1}|\mathbf{x}_{1:T}^{n},\mathbf{y}_{1:T})$, enabling the generation of refined motion at step $n-1$ through sampling. The process concludes at $n=1$, producing the final refined motion $\mathbf{x}_{1:T}$.}
\label{fig:diffusionmodel}
\end{center}
\end{figure}

\noindent\textbf{Reverse Diffusion Process.} The reverse diffusion process characterizes the posterior distribution:
\begin{equation}
    p(\mathbf{x}_{1:T}|\mathbf{y}_{1:T}) = \int p(\mathbf{x}_{1:T},\mathbf{x}_{1:T}^{1:N}|\mathbf{y}_{1:T}) \, \mathrm{d}\mathbf{x}_{1:T}^{1:N},
\end{equation}
where 
\begin{equation}
\resizebox{.98\hsize}{!}{$p(\mathbf{x}_{1:T}^{0:N}|\mathbf{y}_{1:T}) = p(\mathbf{x}_{1:T}^N|\mathbf{y}_{1:T}) \prod_{n=1}^N p_{\mathbf{W}}(\mathbf{x}_{1:T}^{n-1}|\mathbf{x}_{1:T}^{n},\mathbf{y}_{1:T}),
$}\label{eq:reversediffusion}
\end{equation}
with $\mathbf{x}_{1:T}^{0}=\mathbf{x}_{1:T}$ for simplicity in notation. In Equation~\ref{eq:reversediffusion}, $p(\mathbf{x}_{1:T}^N|\mathbf{y}_{1:T}) \approx \mathcal{N}(\mathbf{y}_{1:T},\kappa^2\mathbf{I})$, i.e, a multivariate Gaussian distribution centered at $\mathbf{y}_{1:T}$. $p_{\mathbf{W}}(\mathbf{x}_{1:T}^{n-1}|\mathbf{x}_{1:T}^{n},\mathbf{y}_{1:T})$ represents the learned distribution, with parameters $\mathbf{W}$, to approximate the data distribution $q(\mathbf{x}_{1:T}^{n-1}|\mathbf{x}_{1:T}^{n},\mathbf{x}_{1:T},\mathbf{y}_{1:T})$. Given $p_{\mathbf{W}}(\mathbf{x}_{1:T}^{n-1}|\mathbf{x}_{1:T}^{n},\mathbf{y}_{1:T})$, we can transform $\mathbf{x}_{1:T}^{n}$ to $\mathbf{x}_{1:T}^{n-1}$ and infer the refined motion $\mathbf{x}_{1:T}$ from the initial estimates $\mathbf{y}_{1:T}$ by iteratively denoising from $\mathbf{y}_{1:T}$ towards $\mathbf{x}_{1:T}$ through sampling. In the following, we introduce our framework for capturing the inverse transition distribution $p_{\mathbf{W}}(\mathbf{x}_{1:T}^{n-1}|\mathbf{x}_{1:T}^{n},\mathbf{y}_{1:T})$ in the reverse diffusion process.

To capture $p_{\mathbf{W}}(\mathbf{x}_{1:T}^{n-1}|\mathbf{x}_{1:T}^{n},\mathbf{y}_{1:T})$, prior methods have focused either on predicting the transition noise~\cite{ho2020denoising} or directly estimating the final clean state~\cite{tevet2023human,lee2024interhandgen}. Inspired by the recent success of human diffusion models~\cite{tevet2023human}, we adopt the latter approach. As depicted in Figure~\ref{fig:diffusionmodel}, our framework takes as input the initial motion estimates $\mathbf{y}_{1:T}$ and the refined estimates $\mathbf{x}_{1:T}^n$ at diffusion step $n$, and outputs an estimated clean motion $\hat{\mathbf{x}}_{1:T}=f_\mathbf{W}(\mathbf{x}_{1:T}^n,\mathbf{y}_{1:T},n)$ to parameterize the transition distribution $p_{\mathbf{W}}(\mathbf{x}_{1:T}^{n-1}|\mathbf{x}_{1:T}^{n},\mathbf{y}_{1:T})$. We then generate $\mathbf{x}_{1:T}^{n-1}$ through sampling from $p_{\mathbf{W}}(\mathbf{x}_{1:T}^{n-1}|\mathbf{x}_{1:T}^{n},\mathbf{y}_{1:T})$ and obtain the final refined motion estimates $\mathbf{x}_{1:T}=\hat{\mathbf{x}}_{1:T}$ when $n=1$.

The core of our reverse diffusion framework is a Transformer architecture~\cite{vaswani2017attention}, which effectively captures temporal dependencies across the motion sequence. For feature embedding at each time frame, we first apply forward kinematics to the pose and shape parameters to obtain 3D hand meshes for both the initial and refined motions, denoted as $\mathbf{M}_{1:T}^y$ and $\mathbf{M}_{1:T}^n$, respectively. A mesh convolutional neural network (MeshCNN)~\cite{chen2022mobrecon} is then employed to capture the spatial dependencies within the 3D hand geometry. Additionally, we encode the diffusion step $n$ using a Multi-layer Perceptron (MLP). These input features, combined with temporal positional encoding, are processed by the Transformer to generate clean motion estimates, $\hat{\mathbf{x}}_{1:T}$, through autoregressive prediction. Specifically, the prediction for time step $t+1$ is conditioned on the input data and historical predictions from time steps $1$ to $t$.

To train our framework, given $\mathbf{y}_{1:T}$ and the sample  $\mathbf{x}_{1:T}$ drawn from $q(\mathbf{x}_{1:T}|\mathbf{y}_{1:T})$, we minimize:
\begin{equation}
    \mathcal{L}_{data}= \mathbb{E}_{n\sim[1,N]} \left\| \mathbf{x}_{1:T} - f_\mathbf{W}(\mathbf{x}_{1:T}^n, \mathbf{y}_{1:T}, n) \right\|^2,
    \label{eq:dataloss}
\end{equation}
where $\mathbf{W}$ denotes the learnable parameters in our reverse diffusion framework. Equation~\ref{eq:dataloss} is a well-established objective for diffusion model training~\cite{tevet2023human}. While this approach is effective, such a training strategy is purely data-driven and remains agnostic to how humans efficiently move in different scenarios. Instead, we incorporate intuitive physics of hand motion to improve the model's performance, as detailed in the next section.

\subsection{Improving Hand Dynamics Modeling with Intuitive Physics }
\label{sec:intuitivephysics}

Human hands interact with the physical world based on an intuitive understanding of the scene, moving with remarkable efficiency and accuracy after years of experience. In this section, we study intuitive physics in hand motion and present effective methods for integrating this knowledge into the diffusion model to enhance its performance.

\begin{figure}[t]
\begin{center}
\includegraphics[width=0.99\linewidth]{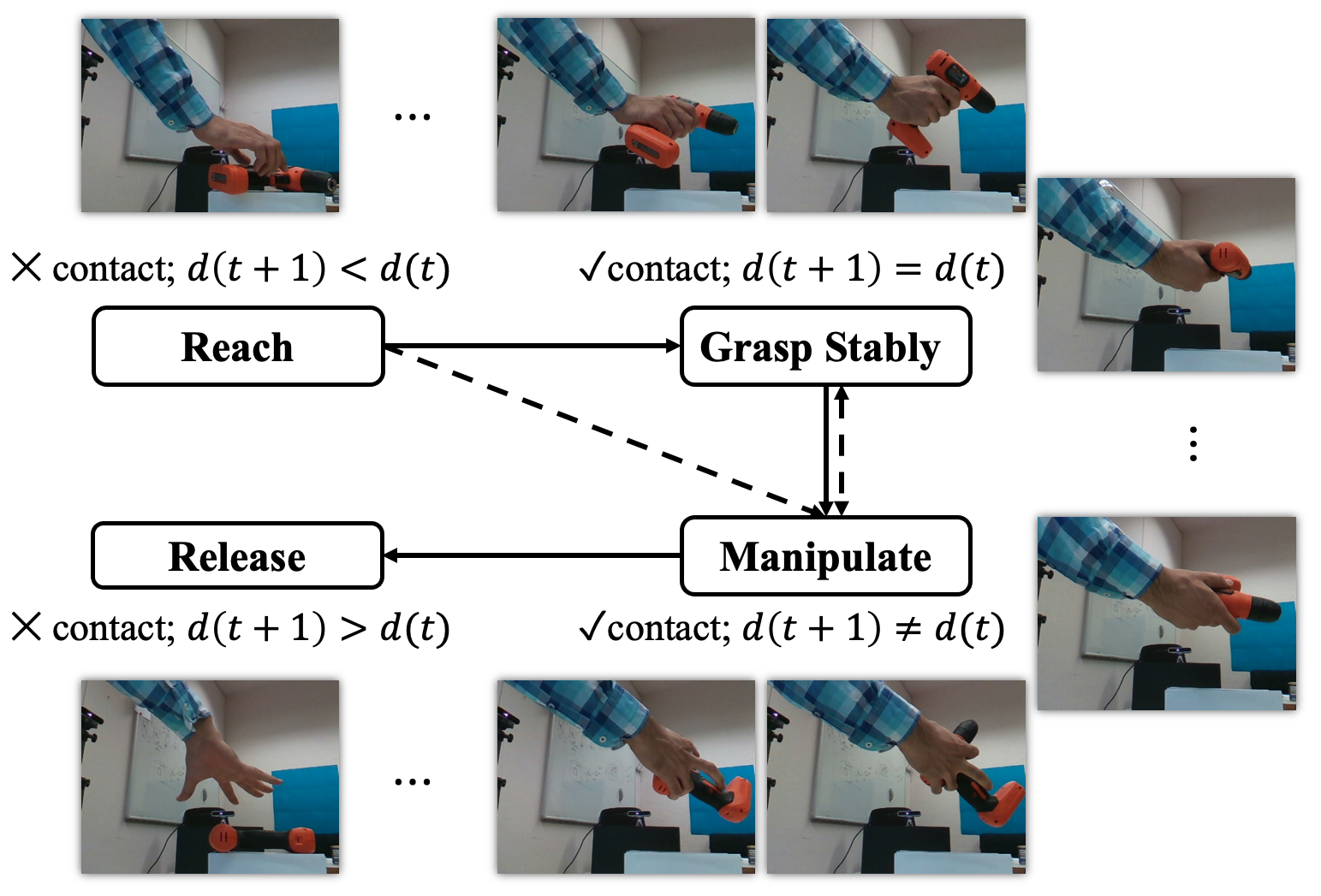}
\caption{\textbf{Illustration of Key Hand Motion States.} Each state is defined by its contact condition and temporal change in hand-object distance ($d(t)$). The contact status indicates whether the hand is touching the object, while distance changes reflect the temporal evolution of hand-object separation. The images are from HO3D~\cite{hampali2020honnotate}. The solid arrow describes the state transition for this example video sequence, while the dashed arrow indicates possible transitions among the four motion states.}
\label{fig:motionstate}
\end{center}
\end{figure}

\noindent\textbf{Identification of Hand Motion Physical Principles.} Hand movements can broadly be categorized into two scenarios: free movements and interactions with external objects. While hand-object interactions are inherently more complex, they follow fundamental principles informed by intuitive physics. As illustrated in Figure~\ref{fig:motionstate}, hand motion during object interaction typically progresses through several well-defined stages. Initially, the hand moves from an idle or free-motion state directly toward the object without unnecessary deviation. Upon contact, the hand grasps the object and holds it in a stable pose. During manipulation, the hand undergoes significant pose changes as it adjusts to various object states. Finally, after completing the manipulation, the hand releases the object and returns to a free-motion state.

Based on these observations, we categorize hand motions into four fundamental states: \textit{reaching}, \textit{stable grasping}, \textit{manipulation}, and \textit{releasing}. Each state is defined by its contact condition and temporal changes in hand-object distance as highlighted in Figure~\ref{fig:motionstate}. Building upon these motion states, we further identify two fundamental physical constraints that characterize efficient and realistic hand motions. Specifically, during reaching and releasing, the hand motion complies with the kinetics constraints: \textit{the hand follows a minimal-effort trajectory either toward or away from the object}. While grasping stably, the hand follows the stability constraints: \textit{the digits stay mostly static}. 

The derived intuitive physics offer straightforward constraints for hand motion but are challenging to capture computationally \cite{garrido2025intuitive}. Our experiments in Section~\ref{sec:exp} also illustrate the limitations of existing 3D hand reconstruction models in enforcing these principles. To address this, we integrate these insights into our diffusion model for enhanced dynamics modeling, as detailed in the following section.

\noindent\textbf{Integration of Intuitive Physics with the Diffusion Model.} To enable the model to learn state-specific dynamics, our approach first augments the 3D motion data with motion state annotations. Specifically, we annotate the motion states for all training sequences and enhance the diffusion-based refinement model to predict motion states for each frame, denoted as $\hat{c}_{1:T}$. We introduce a loss term that minimizes the discrepancy between predicted and annotated states:
\begin{equation}
\resizebox{.5\hsize}{!}{$
    \mathcal{L}_{state}= \frac{1}{T}\sum_{t=1}^T CE(c_{t},\hat{c}_{t}),
$}
    \label{ch6-eq:stateloss}
\end{equation}
where $CE(\cdot,\cdot)$ represents the cross-entropy loss computed using the predicted and annotated motion states for each time frame $\hat{c}_{t}$ and $c_{t}$. By integrating the motion state into the diffusion model, the state modeling serves as a conditioning variable for predicting future motion and states in the autoregressive prediction, as illustrated in the top part of Figure~\ref{fig:diffusionmodel}. Since the predicted states $\hat{c}_{t}$ are discrete one-hot encodings, we employ a Gumbel-Softmax layer~\cite{jang2016categorical} to enable differentiable training.

For the kinetics constraints, we introduce a least kinetic motion constraint that encourages direct paths to target positions and minimizes unnecessary deviations, formulated as:
\begin{equation}
\scalebox{0.9}{$
    \mathcal{L}_{kinetics}= \frac{1}{|\mathcal{C}_{r}|} \sum_{t,t+1,t+2 \in \mathcal{C}_{r}} \max(0, \phi(\boldsymbol{\theta}_{t},\boldsymbol{\theta}_{t+1},\boldsymbol{\theta}_{t+2}))
    $},
    \label{ch6-eq:kinetic}
\end{equation}
where $\mathcal{C}_r$ denotes a set of frame indices belonging to the reaching or releasing states with size $|\mathcal{C}_r|$, $\phi(\boldsymbol{\theta}_{t},\boldsymbol{\theta}_{t+1},\boldsymbol{\theta}_{t+2})=-\text{sign}(\boldsymbol{\theta}_{t}-\boldsymbol{\theta}_{t+1}) (\boldsymbol{\theta}_{t+2} - \boldsymbol{\theta}_{t+3})$ measures the direction change of hand pose, indicating whether the motion continues in the same direction or reverses over three consecutive frames $t,t+1,t+2$. The loss function encourages the predicted hand trajectories to follow minimal-effort paths during reaching and releasing states. 

Moreover, we encode the stability constraint as:
\begin{equation}
\mathcal{L}_{stability}= \frac{1}{|\mathcal{C}_{g}|}\sum_{t,t+1\in \mathcal{C}_g} \|\boldsymbol{\theta}_{f,t}-\boldsymbol{\theta}_{f,t+1}\|_2^2,
\label{ch6-eq:stability}
\end{equation}
where $\mathcal{C}_g$ denotes the set of frame indices belonging to the stable grasping state with size $|\mathcal{C}_g|$, $t$ and $t+1$ are neighboring frames, $\|\cdot\|_2^2$ denotes the squared L2 norm, and $\boldsymbol{\theta}_{f,t}$ represents the finger joint poses at a certain time frame (excluding the root joint). This loss function penalizes variations in hand pose during stable grasping.

\subsection{Overall Model Training and Inference}

Assembling the data-driven term with losses derived from intuitive physics, we define the total training objective as:
\begin{equation}
    \mathcal{L}_{total}= \mathcal{L}_{data} + \lambda_1 \mathcal{L}_{state} + \lambda_2 \mathcal{L}_{kinetic}+\lambda_3\mathcal{L}_{stability}.
    \label{eq:kineticloss}
\end{equation}
During training, we follow \cite{baradel2022posebert,zhang2024physpt} by using randomly perturbed ground-truth motion as initial motion estimates, enabling our model to refine predictions from any frame-wise reconstruction method. In Appendix~\ref{appx:fullysupervised}, we demonstrate more substantial improvements when applied to a specific frame-wise reconstruction method. At inference time, given initial estimates from any frame-wise prediction model, we sample from the posterior distribution defined in Equation~\ref{eq:reversediffusion} to iteratively generate refined motion predictions. The detailed training and inference procedures are provided in Algorithm~\ref{al:training} and Algorithm~\ref{al:inference}, respectively.

\begin{algorithm}
\caption{Model Training}
\begin{algorithmic}[1]
\Require $\eta_{1:N}$: diffusion noise scheduling; $\kappa$: noise variance level; $q(\mathbf{x}_{1:T},\mathbf{y}_{1:T})$: data distribution
\Repeat
    \State Sample $(\mathbf{x}_{1:T},\mathbf{y}_{1:T})$ from $q(\mathbf{x}_{1:T},\mathbf{y}_{1:T})$ 
        
    \State Compute diffused data at step $n$ (Equation~\ref{eq:forwarddiffusion}) 
    \Statex \hspace{1cm} $\epsilon \sim \mathcal{N}(\mathbf{0}, \mathbf{I})$
    \Statex \hspace{1cm} $\mathbf{e}_{1:T}=\mathbf{y}_{1:T}-\mathbf{x}_{1:T}$
    \Statex \hspace{1cm} $\mathbf{x}_{1:T}^n=\mathbf{x}_{1:T}+\eta_n\mathbf{e}_{1:T}+\kappa\sqrt{\eta_n}\epsilon$
    \State Take a gradient step on $\nabla_{\mathbf{W}} \mathcal{L}_{total}$ (Equation~\ref{eq:kineticloss})
\Until{converged}
\end{algorithmic}
\label{al:training}
\end{algorithm}
\begin{algorithm}
\caption{Model Inference}
\begin{algorithmic}[1]
\Require $\eta_{1:N}$: diffusion noise scheduling; $\kappa$: noise variance level; $f_\mathbf{W}(\mathbf{x}_{1:T}^n, \mathbf{y}_{1:T}, n)$: trained denoising model; $\mathbf{y}_{1:T}$: initial motion estimates

\State $\mathbf{x}_{1:T}^{N} \sim \mathcal{N}(\mathbf{y}_{1:T}, \kappa\mathbf{I})$

\ForAll{$n$ from $N$ to 1}

    \State Estimate the inverse transition distribution $p_{\mathbf{W}}(\mathbf{x}_{1:T}^{n-1}|\mathbf{x}_{1:T}^{n},\mathbf{y}_{1:T})$ via $\hat{\mathbf{x}}_{1:T}=f_\mathbf{W}(\mathbf{x}_{1:T}^n, \mathbf{y}_{1:T}, n)$
    \State Sample $\mathbf{x}_{1:T}^{n-1}$ from $p_{\mathbf{W}}(\mathbf{x}_{1:T}^{n-1}|\mathbf{x}_{1:T}^{n},\mathbf{y}_{1:T})$

\EndFor

\State $\mathbf{x}_{1:T}=\mathbf{x}_{1:T}^{0}$
\end{algorithmic}
\label{al:inference}
\end{algorithm}
\section{Experiment}
\label{sec:exp}
In this section, we validate the effectiveness of our proposed physics-augmented diffusion-based motion refinement model. We begin by outlining the evaluation protocol, including the datasets used, the evaluation metrics employed, and the implementation details in Section~\ref{sec:implementation}. Next, in Section~\ref{sec:sota}, we compare our method with state-of-the-art (SOTA) approaches, demonstrating its enhanced performance. Finally, we present the ablation study in Section~\ref{sec:ablation}, analyzing the contribution of each component of our model. 

\subsection{Datasets, Metrics, and Implementation Details}
\label{sec:implementation}
\noindent\textbf{Datasets.} We conduct our experiments on the widely adopted DexYCB~\cite{chao2021dexycb} and HO3Dv2~\cite{hampali2020honnotate} datasets. Both datasets consist of multi-view data capturing hand-object interactions, often with excessive occlusions. We adhere to the standard training and testing splits of DexYCB for fair comparison with other works. We test our DexYCB-trained model directly on HO3Dv2 without fine-tuning, posing a more challenging setting compared to existing methods.

\noindent\textbf{Evaluation Metrics.} Following the protocol of previous methods~\cite{chao2021dexycb,zhang2024weakly,pavlakos2023reconstructing,kocabas2020vibe}, we evaluate 3D reconstruction error (\textit{Rec. Error}) and physical plausibility (\textit{Phys. Plausibility}). \textit{Rec. Error} includes the average Euclidean distance between the predicted and the ground truth 3D hand joint positions before and after Procrustes alignment (MJE/P-MJE). \textit{Phys. Plausibility} involves the average difference between the predicted and the ground truth acceleration ACCL (acceleration error in mm/frame$^2$). We also compute the violation of the least kinetic constraint defined in Equation~\ref{ch6-eq:kinetic} (KIN in degrees) and the stability constraint defined in Equation~\ref{ch6-eq:stability} (STA in degrees). For the evaluation on HO3Dv2, we report mesh vertex reconstruction error (P-MVE) and the fraction of error below 5\,mm and 15\,mm (F@5 and F@15) following the established evaluation system~\cite{hampali2020honnotate}.

\noindent\textbf{Implementation.} For the model architecture, the Transformer backbone consists of standard encoder and decoder layers, with 4 layers, 8 attention heads, and 512 embedding dimensions. The MeshCNN has four layers with feature dimensions of [32, 64, 64, 64]. Aligning with most existing methods, the input sequence length is 16. For data preprocessing, we apply a Gaussian filter to smooth the data to reduce the impact of unrelated noise introduced during data collection. For model training, we employ the AdamW~\cite{loshchilov2017decoupled} optimizer with an initial learning rate of $10^{-4}$, which decreases by a factor of 0.8 after every 5 epochs. The hyperparameters are empirically set as: $\lambda_{1} = 50$, $\lambda_{2} = 5e^{2}$, and $\lambda_{3} = 1e^{3}$. We also incorporate the geometric losses used in \cite{tevet2023human} to encourage accurate geometric position prediction, as well as the trajectory consistency loss from~\cite{kim2023consistency} to mitigate error accumulation in the diffusion process.

\subsection{Comparison with SOTA Methods}
\label{sec:sota}

In this section, we first highlight the consistent enhancements over SOTA motion refinement models when applied to different frame-wise reconstruction methods. We then showcase our advantages and improved performance over existing video-based 3D hand reconstruction methods.

\begin{table}[t]
\tabcolsep=0.02in
    \begin{center}
    \begin{tabular}{ l cc  ccc }
    \toprule
    \multirow[b]{2}{*}{Method} & \multicolumn{2}{c}{\textit{Rec. Error}} & \multicolumn{3}{c}{\textit{Phys. Plausibility}}  \\ \cmidrule(lr){2-3} \cmidrule{4-6}
     & MJE & P-MJE & ACCL & KIN & STA \\ \cmidrule{1-6}
    A: K-Hand \cite{zhang2024weakly} & 24.4 & 5.8 & 14.41 & 23.24 & 1.68  \\
    B: HaMer \cite{pavlakos2023reconstructing} & 18.9 & 4.4 & 7.95 & 22.49 & 1.08 \\
    \cmidrule{1-6}
    A + PoseBERT \cite{baradel2022posebert} & 22.1/22.3 & 6.0/5.5 & 2.96/3.33 & 0.73/- & 0.00/- \\
    B + PoseBERT \cite{baradel2022posebert} & 18.4/18.0 & 5.0/4.4 & 2.71/2.38 & 0.67/- & 0.00/- \\
    \cmidrule{1-6}
    A + \textbf{Ours} & \textbf{21.5} & \textbf{5.3} & \textbf{1.17} & \textbf{0.00} & \textbf{0.00} \\
    B + \textbf{Ours} & \textbf{17.5} & \textbf{4.1} & \textbf{1.01} & \textbf{0.00} & \textbf{0.00} \\
    \bottomrule
    \end{tabular}
    \caption{\textbf{Comparison with SOTA Motion Refinement on Different Frame-wise Reconstruction Models.} The evaluation is conducted on the DexYCB dataset. We compare our method against the SOTA motion refinement baseline PoseBERT on two frame-wise reconstruction models: (A) K-Hand and (B) HaMer. Results of PoseBERT (\(\cdot/\cdot\)) represent its two versions refining joint angles/positions. For all the values, the smaller, the better.}
    \label{tab:sotarefine}
    \end{center}
\end{table}

\noindent\textbf{Comparison with SOTA Motion Refinement.} The most closely related work to our approach is PoseBERT, a leading motion refinement model that employs a Transformer encoder model trained with random input masking and noise perturbations. We compare our physics-augmented diffusion-based motion refinement model with PoseBERT on two widely used frame-wise reconstruction models: KNOWN-Hand (K-Hand) and HaMer. K-Hand leverages generic hand knowledge, while HaMer relies on large-scale hand data to achieve good reconstruction accuracy. While both PoseBERT and our method improve upon these frame-wise reconstructions, our method achieves more significant improvements across all metrics, in terms of both reconstruction accuracy and physical plausibility as shown in Table~\ref{tab:sotarefine}. Specifically, when refining K-Hand (24.4 mm MJE), our method achieves 21.7 mm MJE, outperforming PoseBERT's 22.3 mm. Similarly for HaMer (18.9 mm MJE), we achieve 17.5 mm MJE, better than PoseBERT's 18.0 mm. The improvements in physical plausibility are even more pronounced. Our method reduces ACCL values to 1.30 mm/frame$^2$ and 1.18 mm/frame$^2$, compared to K-Hand's 14.41 mm/frame$^2$ and HaMer's 7.95 mm/frame$^2$, substantially surpassing PoseBERT's improvements of 3.33 and 2.38 mm/frame$^2$, respectively. These consistent improvements across different frame-wise models demonstrate the effectiveness of our approach in leveraging diffusion modeling and intuitive physics for enhanced motion refinement. Next, we demonstrate our advantages over existing methods for 3D hand reconstruction from monocular videos.

\begin{table}[t]
\tabcolsep=0.02in
    \begin{center}
    \begin{tabular}{ l cc  cc }
    \toprule
    Method & P-MJE & P-MVE & F@5  & F@15  \\ 
    \cmidrule{1-5}
    HaMer \cite{pavlakos2023reconstructing} &8.1 & 8.6 &58.0&97.4\\
    \cmidrule{1-5}
    VIBE~\cite{kocabas2020vibe} & 9.9 & 9.5 & 52.6 & 95.5 \\
    TCMR~\cite{choi2021beyond} & 11.4 & 10.9 & 46.3 & 93.3  \\
    TempCLR~\cite{ziani2022tempclr} & 10.6 & 10.6 & 48.1 & 93.7  \\
    Deformer~\cite{Fu_2023_ICCV}& 9.4 & 9.1 & 54.6 & 96.3 \\    
    \cmidrule{1-5}
     \textbf{Ours}$^*$ & \textbf{8.0} & \textbf{8.3} & \textbf{59.7} & \textbf{97.6} \\
    \bottomrule
    \end{tabular}
    \caption{\textbf{Comparison with SOTA Video-based Reconstruction.} The evaluation is on HO3Dv2. The frame-wise reconstruction we use is HaMer. Results of other methods are obtained from published papers~\cite{Fu_2023_ICCV}. $*$ denotes a zero-shot setting where our method is not trained on HO3Dv2. Lower values are better for P-MJE and P-MVE, while higher values are better for F@5 and F@15.}
    \label{tab:sota}
    \end{center}
\end{table}

\noindent\textbf{Comparison with SOTA Video-based Methods.} Existing methods for reconstructing 3D hand motion from monocular videos typically require training on annotated video sequences, which limits their performance due to the scarcity of such data. In contrast, our approach solely exploits 3D motion data and effectively leverages diffusion models and physics principles to achieve superior performance. Applying our model on top of the leading frame-wise reconstruction model yields the best performance for 3D hand reconstruction from videos. As shown in Table~\ref{tab:sota}, our method significantly outperforms existing video-based methods on the HO3Dv2 dataset, achieving the best results across all metrics (P-MJE: 8.0 mm, P-MVE: 8.3 mm, F@5: 59.7\%, F@15: 97.6\%). Notably, we achieve these improvements without training on HO3Dv2 but solely on DexYCB, highlighting the generalizability of our physics-augmented diffusion model for hand motion refinement. In Appendix~\ref{appx:dynamics}, we further present consistent improvements on complex bimanual hand-object manipulation data.

\begin{figure}[t]
\begin{center}
\includegraphics[width=0.99\linewidth]{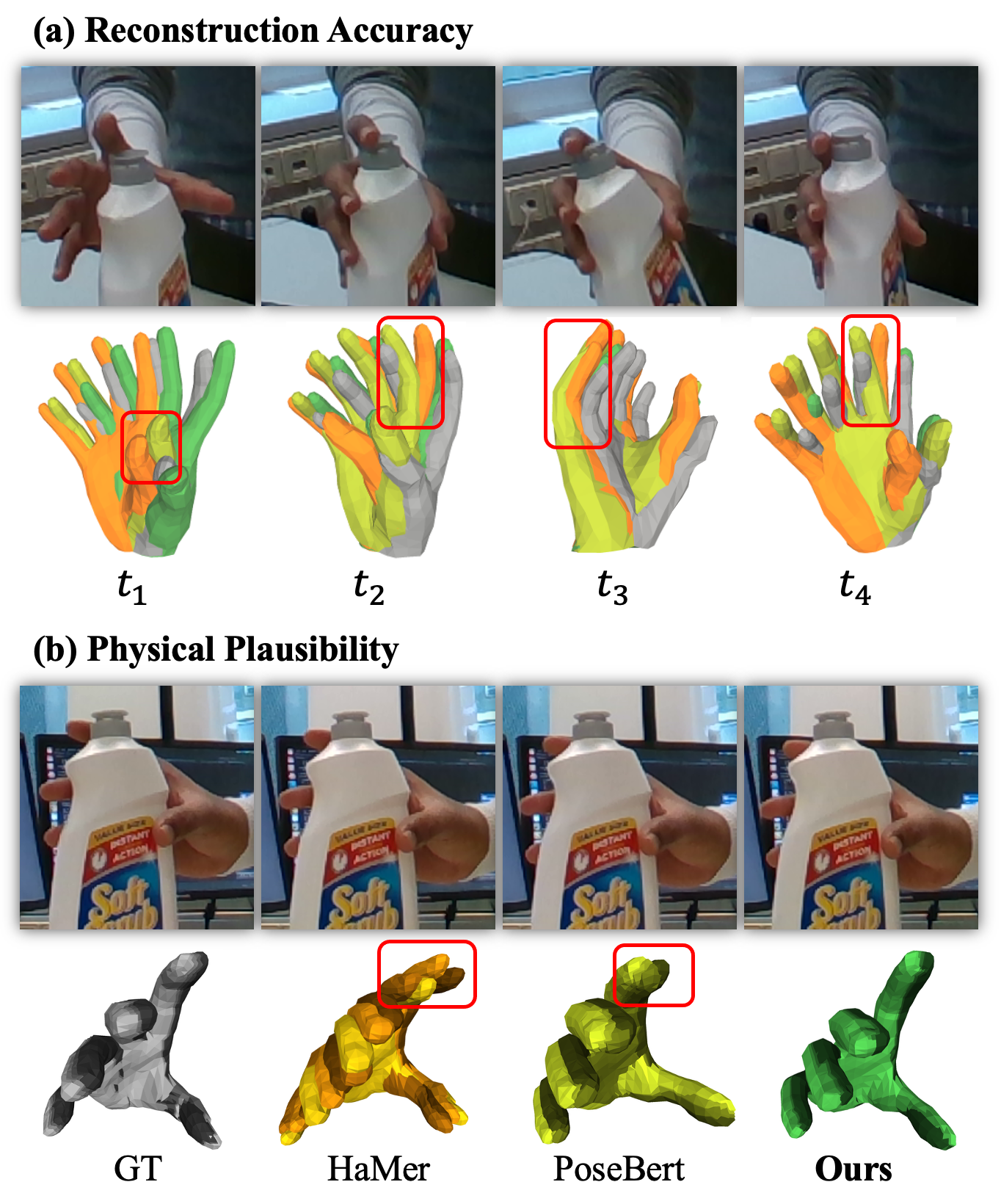}
\vspace{-0.2cm}
\caption{\textbf{Qualitative Evaluation.} The test sequences are from HO3Dv2, with the video frames presented in causal time order from left to right. The hand reconstructions are from the ground truth (GT, gray), the leading frame-wise reconstruction model (HaMer, orange), the leading motion refinement model (PoseBERT, lime), and our method (Ours, green). The red boxes mark degraded reconstructions from HaMer and PoseBERT. (a) shows comparisons with GT across time frames $t_{1}$ to $t_{4}$. (b) illustrates motion consistency, where lighter colors indicate later frames. Additionally, we highlight our advantages in handling stable grasping sequences through more qualitative evaluations in Appendix~\ref{appx:additionalqualitive}.}
\label{fig:addqualitative}
\end{center}
\end{figure}

\noindent\textbf{Qualitative Evaluation.} To further validate our approach, we provide qualitative evaluation in Figure~\ref{fig:addqualitative}. Our method consistently outperforms SOTA 3D hand reconstruction and motion refinement models in both reconstruction accuracy and physical plausibility. For the reaching and grasping of a bottle testing sequence in Figure~\ref{fig:addqualitative}(a), the frame-wise reconstruction model HaMer produces inaccurate 3D hand configurations, and the motion refinement model PoseBERT fails to correct these errors, even when leveraging temporal information. In the testing sequence involving subtle hand motions (Figure~\ref{fig:addqualitative}(b)), PoseBERT exhibits improved temporal smoothness over HaMer, but the results still fall short of the coherent ground-truth motion. In contrast, our method achieves the best reconstruction quality and demonstrates strong physical plausibility in these hand-object interaction examples, highlighting the effectiveness of our physics-augmented diffusion-based framework.

\subsection{Ablation Study}
\label{sec:ablation}

\begin{table}[t]
\tabcolsep=0.02in
    \begin{center}
    \begin{tabular}{ c ccc  cc  ccc }
    \toprule
    Model & \multicolumn{3}{c}{Physics Losses} & \multicolumn{2}{c}{\textit{Rec. Error}} & \multicolumn{3}{c}{\textit{Phys. Plausibility}}  \\ \cmidrule{1-1} \cmidrule(lr){2-4} \cmidrule{5-6} \cmidrule(lr){7-9}
    \textit{Prob.} & $\mathcal{L}_{state}$ & $\mathcal{L}_{kin}$ & $\mathcal{L}_{sta}$ &MJE & P-MJE & ACCL & KIN & STA \\ \cmidrule{1-9}
     \multicolumn{4}{c}{Frame-wise Rec.} & 24.4 & 5.8 & 14.41 & 23.24 & 1.68 \\ \cmidrule{1-9}
    \multicolumn{4}{c}{SmoothFilter~\cite{young1995recursive}} & 25.3 & 5.4 & 1.96 & 0.00 & 0.00 \\
    \multicolumn{4}{c}{Constant Accl. Loss} & 21.7 & 5.5 & 1.29 & 0.00 & 0.00\\ 
    \cmidrule{1-9}
     &  & & & 22.7 & 5.3 & 2.53 & 0.39 & 0.04 \\
    $\checkmark$ &  & & & 21.9 & 5.3 & 1.38 & 0.12 & 0.02 \\
    $\checkmark$ & $\checkmark$ & & & 21.6 & 5.3 & 1.27 & 0.06 & 0.00  \\
    $\checkmark$ & $\checkmark$ & $\checkmark$ & & 21.7 & 5.3 & 1.17 & 0.00 & 0.00 \\
    $\checkmark$ & $\checkmark$ & & $\checkmark$ & 21.5 & 5.3 & 1.27 & 0.05 & 0.00 \\
    \cmidrule{1-9}
    $\checkmark$ & $\checkmark$ & $\checkmark$ & $\checkmark$ & \textbf{21.5} & \textbf{5.3} & \textbf{1.17} & \textbf{0.00} & \textbf{0.00} \\
    \bottomrule
    \end{tabular}
    \caption{\noindent\textbf{Ablation on the Diffusion Model and the Training Losses Derived from Intuitive Physics.} The evaluation is conducted on DexYCB.  ``SmoothFilter" and ``Constant Accl. Loss" are heuristic methods to promote smoothness that apply a low-pass filter to the estimates or enforce constant acceleration during training (like \cite{christen2022d}), respectively. \textit{Prob.} indicates whether the model is probabilistic. The deterministic model is a Transformer, while being probabilistic means using our proposed diffusion-based motion refinement. $\mathcal{L}_{kin}$ and $\mathcal{L}_{sta}$ denotes the kinetic and stability constraint, respectively. For all the values, the smaller, the better.}
    \label{tab:ablation}
    \end{center}
\end{table}

This sections demonstrates the effectiveness of each component of our method, including modeling the distribution of refined motion estimates conditioned on the initial ones through diffusion models and integrating dynamic intuitive physics. We present the evaluation results in Table~\ref{tab:ablation}, where we employ K-Hand~\cite{zhang2024weakly} for the frame-wise reconstruction.

\noindent\textbf{Diffusion-based Model.} As shown in the table, we first evaluate a baseline approach that employs a Transformer to learn a deterministic mapping from initial to refined motion estimates, without leveraging diffusion modeling. While this approach improves on frame-wise evaluation by capturing temporal dependencies, it fails to capture the inherent uncertainty in the initial motion estimates. Instead, our proposed diffusion-based motion refinement framework achieves further improvements in both reconstruction accuracy and physical plausibility. Specifically, the MJE decreases from 22.7 to 21.9 mm, while the ACCL shows a more substantial 45\% improvement, dropping from 2.53 to 1.38 mm/frame$^2$. These improvements demonstrate the effectiveness of our approach in handling the uncertainty through an iterative denoising process.

\noindent\textbf{Integration of Intuitive Physics.} Purely data-driven models remain agnostic to fundamental physics principles, yielding inferior reconstruction accuracy and struggling with physical plausibility. As shown in Table~\ref{tab:ablation}, incorporating the four key hand motion states into the diffusion model improves both reconstruction accuracy and physical plausibility. Specifically, the MJE improves from 21.9 to 21.6 mm, and the ACCL improves from 1.38 to 1.27 mm/frame$^2$. Furthermore, incorporating the loss derived from the least kinetic motion constraint and the stability loss effectively enhances physical plausibility. For example, utilizing the least kinetic motion constraint reduces the ACCL further to 1.17 mm/frame$^2$ and minimizing the violations of both kinetic and stability constraints to nearly zero. 

Finally, we achieve the best model performance by leveraging both the diffusion model and the intuitive physics. It is worth noting that heuristic methods do not yield comparable results. Specifically, smoothing predictions with a Gaussian filter (``SmoothFilter") may produce smoother trajectories with improved physical plausibility, but their reconstruction accuracy tends to suffer. Moreover, assuming a constant acceleration of predicted motions (``Constant Accl. Loss") leads to suboptimal results. In contrast, our approach achieves better performance in both reconstruction accuracy and physical plausibility. It more effectively captures the unique behaviors at different motion states compared to the assumption of constant acceleration.

\section{Conclusion}

In this work, we present a novel method for estimating 3D hand motion from monocular videos by integrating diffusion models with intuitive hand physics. We introduce a dedicated diffusion-based framework that refines frame-wise 3D hand reconstructions through iterative denoising, capturing the distribution of refined estimates conditioned on the initial predictions. Moreover, we identify valuable intuitive physics knowledge related to hand motions. We effectively incorporate them into the diffusion model through a state modeling loss and state-specific constraints. Our method consistently outperforms existing motion refinement models. When applied to leading frame-wise reconstruction methods, it achieves state-of-the-art performance in 3D hand motion estimation, improving both reconstruction accuracy and physical plausibility. Notably, it addresses a key limitation of purely data-driven models, which can fail even during simple static motions—such as stable object grasping—when the visual input is degraded.

\section*{Acknowledgment}
This work is supported in part by IBM through the IBM-Rensselaer Future of Computing Research Collaboration and in part by the DARPA grant HR00112390059.

{
    \small
    \bibliographystyle{ieeenat_fullname}
    \bibliography{main}

\begin{thebibliography}{84}
\providecommand{\natexlab}[1]{#1}
\providecommand{\url}[1]{\texttt{#1}}
\expandafter\ifx\csname urlstyle\endcsname\relax
  \providecommand{\doi}[1]{doi: #1}\else
  \providecommand{\doi}{doi: \begingroup \urlstyle{rm}\Url}\fi

\bibitem[Baek et~al.(2019)Baek, Kim, and Kim]{baek2019pushing}
Seungryul Baek, Kwang~In Kim, and Tae-Kyun Kim.
\newblock Pushing the envelope for rgb-based dense 3d hand pose estimation via neural rendering.
\newblock In \emph{Proceedings of the IEEE/CVF Conference on Computer Vision and Pattern Recognition}, pages 1067--1076, 2019.

\bibitem[Baek et~al.(2020)Baek, Kim, and Kim]{baek2020weakly}
Seungryul Baek, Kwang~In Kim, and Tae-Kyun Kim.
\newblock Weakly-supervised domain adaptation via gan and mesh model for estimating 3d hand poses interacting objects.
\newblock In \emph{Proceedings of the IEEE/CVF Conference on Computer Vision and Pattern Recognition}, pages 6121--6131, 2020.

\bibitem[Bai et~al.(2020)Bai, Sasikumar, Yang, and Billinghurst]{bai2020user}
Huidong Bai, Prasanth Sasikumar, Jing Yang, and Mark Billinghurst.
\newblock A user study on mixed reality remote collaboration with eye gaze and hand gesture sharing.
\newblock In \emph{Proceedings of the 2020 CHI conference on human factors in computing systems}, pages 1--13, 2020.

\bibitem[Baradel et~al.(2022)Baradel, Br{\'e}gier, Groueix, Weinzaepfel, Kalantidis, and Rogez]{baradel2022posebert}
Fabien Baradel, Romain Br{\'e}gier, Thibault Groueix, Philippe Weinzaepfel, Yannis Kalantidis, and Gr{\'e}gory Rogez.
\newblock Posebert: A generic transformer module for temporal 3d human modeling.
\newblock \emph{IEEE Transactions on Pattern Analysis and Machine Intelligence}, 45\penalty0 (11):\penalty0 12798--12815, 2022.

\bibitem[Bhatt and Varadhan(2017)]{bhatt2017hand}
Nayan Bhatt and SKM Varadhan.
\newblock Hand posture comparison in synergy space.
\newblock \emph{PeerJ PrePrints}, 2017.

\bibitem[Boukhayma et~al.(2019)Boukhayma, Bem, and Torr]{boukhayma20193d}
Adnane Boukhayma, Rodrigo~de Bem, and Philip~HS Torr.
\newblock 3d hand shape and pose from images in the wild.
\newblock In \emph{Proceedings of the IEEE/CVF Conference on Computer Vision and Pattern Recognition}, pages 10843--10852, 2019.

\bibitem[Chao et~al.(2021)Chao, Yang, Xiang, Molchanov, Handa, Tremblay, Narang, Van~Wyk, Iqbal, Birchfield, et~al.]{chao2021dexycb}
Yu-Wei Chao, Wei Yang, Yu Xiang, Pavlo Molchanov, Ankur Handa, Jonathan Tremblay, Yashraj~S Narang, Karl Van~Wyk, Umar Iqbal, Stan Birchfield, et~al.
\newblock Dexycb: A benchmark for capturing hand grasping of objects.
\newblock In \emph{Proceedings of the IEEE/CVF Conference on Computer Vision and Pattern Recognition}, pages 9044--9053, 2021.

\bibitem[Chen et~al.(2023)Chen, Yang, and Yao]{chen2023mhentropy}
Rongyu Chen, Linlin Yang, and Angela Yao.
\newblock Mhentropy: Entropy meets multiple hypotheses for pose and shape recovery.
\newblock In \emph{Proceedings of the IEEE/CVF International Conference on Computer Vision}, pages 14840--14849, 2023.

\bibitem[Chen et~al.(2022)Chen, Liu, Dong, Zhang, Ma, Xiong, Zhang, and Guo]{chen2022mobrecon}
Xingyu Chen, Yufeng Liu, Yajiao Dong, Xiong Zhang, Chongyang Ma, Yanmin Xiong, Yuan Zhang, and Xiaoyan Guo.
\newblock Mobrecon: Mobile-friendly hand mesh reconstruction from monocular image.
\newblock In \emph{Proceedings of the IEEE/CVF Conference on Computer Vision and Pattern Recognition}, pages 20544--20554, 2022.

\bibitem[Chen et~al.(2021)Chen, Tu, Kang, Bao, Zhang, Zhe, Chen, and Yuan]{chen2021model}
Yujin Chen, Zhigang Tu, Di Kang, Linchao Bao, Ying Zhang, Xuefei Zhe, Ruizhi Chen, and Junsong Yuan.
\newblock Model-based 3d hand reconstruction via self-supervised learning.
\newblock In \emph{Proceedings of the IEEE/CVF Conference on Computer Vision and Pattern Recognition}, pages 10451--10460, 2021.

\bibitem[Choi et~al.(2020)Choi, Moon, and Lee]{choi2020pose2mesh}
Hongsuk Choi, Gyeongsik Moon, and Kyoung~Mu Lee.
\newblock Pose2mesh: Graph convolutional network for 3d human pose and mesh recovery from a 2d human pose.
\newblock In \emph{European Conference on Computer Vision}, pages 769--787. Springer, 2020.

\bibitem[Choi et~al.(2021)Choi, Moon, Chang, and Lee]{choi2021beyond}
Hongsuk Choi, Gyeongsik Moon, Ju~Yong Chang, and Kyoung~Mu Lee.
\newblock Beyond static features for temporally consistent 3d human pose and shape from a video.
\newblock In \emph{Proceedings of the IEEE/CVF conference on computer vision and pattern recognition}, pages 1964--1973, 2021.

\bibitem[Christen et~al.(2022)Christen, Kocabas, Aksan, Hwangbo, Song, and Hilliges]{christen2022d}
Sammy Christen, Muhammed Kocabas, Emre Aksan, Jemin Hwangbo, Jie Song, and Otmar Hilliges.
\newblock D-grasp: Physically plausible dynamic grasp synthesis for hand-object interactions.
\newblock In \emph{Proceedings of the IEEE/CVF Conference on Computer Vision and Pattern Recognition}, pages 20577--20586, 2022.

\bibitem[Dong et~al.(2024)Dong, Chharia, Gou, Carrasco, and De~la Torre]{dong2024hamba}
Haoye Dong, Aviral Chharia, Wenbo Gou, Francisco~Vicente Carrasco, and Fernando De~la Torre.
\newblock Hamba: Single-view 3d hand reconstruction with graph-guided bi-scanning mamba.
\newblock \emph{arXiv preprint arXiv:2407.09646}, 2024.

\bibitem[Duran et~al.(2024)Duran, Kocabas, Choutas, Fan, and Black]{duran2024hmp}
Enes Duran, Muhammed Kocabas, Vasileios Choutas, Zicong Fan, and Michael~J Black.
\newblock Hmp: Hand motion priors for pose and shape estimation from video.
\newblock In \emph{Proceedings of the IEEE/CVF Winter Conference on Applications of Computer Vision}, pages 6353--6363, 2024.

\bibitem[Fu et~al.(2023)Fu, Liu, Xu, Niebles, and Kitani]{Fu_2023_ICCV}
Qichen Fu, Xingyu Liu, Ran Xu, Juan~Carlos Niebles, and Kris~M. Kitani.
\newblock Deformer: Dynamic fusion transformer for robust hand pose estimation.
\newblock In \emph{Proceedings of the IEEE/CVF International Conference on Computer Vision (ICCV)}, pages 23600--23611, 2023.

\bibitem[Gao et~al.(2022)Gao, Zhang, Chen, Tan, Zhang, Pan, and Tan]{gao2022cyclehand}
Daiheng Gao, Xindi Zhang, Xingyu Chen, Andong Tan, Bang Zhang, Pan Pan, and Ping Tan.
\newblock Cyclehand: Increasing 3d pose estimation ability on in-the-wild monocular image through cyclic flow.
\newblock In \emph{Proceedings of the 30th ACM International Conference on Multimedia}, pages 2452--2463, 2022.

\bibitem[Garrido et~al.(2025)Garrido, Ballas, Assran, Bardes, Najman, Rabbat, Dupoux, and LeCun]{garrido2025intuitive}
Quentin Garrido, Nicolas Ballas, Mahmoud Assran, Adrien Bardes, Laurent Najman, Michael Rabbat, Emmanuel Dupoux, and Yann LeCun.
\newblock Intuitive physics understanding emerges from self-supervised pretraining on natural videos.
\newblock \emph{arXiv preprint arXiv:2502.11831}, 2025.

\bibitem[Ge et~al.(2019)Ge, Ren, Li, Xue, Wang, Cai, and Yuan]{ge20193d}
Liuhao Ge, Zhou Ren, Yuncheng Li, Zehao Xue, Yingying Wang, Jianfei Cai, and Junsong Yuan.
\newblock 3d hand shape and pose estimation from a single rgb image.
\newblock In \emph{Proceedings of the IEEE/CVF Conference on Computer Vision and Pattern Recognition}, pages 10833--10842, 2019.

\bibitem[Grubert et~al.(2018)Grubert, Witzani, Ofek, Pahud, Kranz, and Kristensson]{grubert2018effects}
Jens Grubert, Lukas Witzani, Eyal Ofek, Michel Pahud, Matthias Kranz, and Per~Ola Kristensson.
\newblock Effects of hand representations for typing in virtual reality.
\newblock In \emph{2018 IEEE Conference on Virtual Reality and 3D User Interfaces (VR)}, pages 151--158. IEEE, 2018.

\bibitem[Hampali et~al.(2020)Hampali, Rad, Oberweger, and Lepetit]{hampali2020honnotate}
Shreyas Hampali, Mahdi Rad, Markus Oberweger, and Vincent Lepetit.
\newblock Honnotate: A method for 3d annotation of hand and object poses.
\newblock In \emph{CVPR}, 2020.

\bibitem[Hao et~al.(2024)Hao, Zhang, Zhuo, Wen, and Fan]{hao2024hand}
Yuze Hao, Jianrong Zhang, Tao Zhuo, Fuan Wen, and Hehe Fan.
\newblock Hand-centric motion refinement for 3d hand-object interaction via hierarchical spatial-temporal modeling.
\newblock In \emph{Proceedings of the AAAI Conference on Artificial Intelligence}, pages 2076--2084, 2024.

\bibitem[Hasson et~al.(2019)Hasson, Varol, Tzionas, Kalevatykh, Black, Laptev, and Schmid]{hasson2019learning}
Yana Hasson, Gul Varol, Dimitrios Tzionas, Igor Kalevatykh, Michael~J Black, Ivan Laptev, and Cordelia Schmid.
\newblock Learning joint reconstruction of hands and manipulated objects.
\newblock In \emph{Proceedings of the IEEE/CVF conference on computer vision and pattern recognition}, pages 11807--11816, 2019.

\bibitem[Hasson et~al.(2020)Hasson, Tekin, Bogo, Laptev, Pollefeys, and Schmid]{hasson2020leveraging}
Yana Hasson, Bugra Tekin, Federica Bogo, Ivan Laptev, Marc Pollefeys, and Cordelia Schmid.
\newblock Leveraging photometric consistency over time for sparsely supervised hand-object reconstruction.
\newblock In \emph{Proceedings of the IEEE/CVF conference on computer vision and pattern recognition}, pages 571--580, 2020.

\bibitem[Ho et~al.(2020)Ho, Jain, and Abbeel]{ho2020denoising}
Jonathan Ho, Ajay Jain, and Pieter Abbeel.
\newblock Denoising diffusion probabilistic models.
\newblock \emph{Advances in neural information processing systems}, 33:\penalty0 6840--6851, 2020.

\bibitem[Jang et~al.(2016)Jang, Gu, and Poole]{jang2016categorical}
Eric Jang, Shixiang Gu, and Ben Poole.
\newblock Categorical reparameterization with gumbel-softmax.
\newblock \emph{arXiv preprint arXiv:1611.01144}, 2016.

\bibitem[Jiang et~al.(2023)Jiang, Rahmani, Black, and Williams]{jiang2023probabilistic}
Zheheng Jiang, Hossein Rahmani, Sue Black, and Bryan~M Williams.
\newblock A probabilistic attention model with occlusion-aware texture regression for 3d hand reconstruction from a single rgb image.
\newblock In \emph{Proceedings of the IEEE/CVF Conference on Computer Vision and Pattern Recognition}, pages 758--767, 2023.

\bibitem[Kim et~al.(2023)Kim, Lai, Liao, Murata, Takida, Uesaka, He, Mitsufuji, and Ermon]{kim2023consistency}
Dongjun Kim, Chieh-Hsin Lai, Wei-Hsiang Liao, Naoki Murata, Yuhta Takida, Toshimitsu Uesaka, Yutong He, Yuki Mitsufuji, and Stefano Ermon.
\newblock Consistency trajectory models: Learning probability flow ode trajectory of diffusion.
\newblock \emph{arXiv preprint arXiv:2310.02279}, 2023.

\bibitem[Kocabas et~al.(2020)Kocabas, Athanasiou, and Black]{kocabas2020vibe}
Muhammed Kocabas, Nikos Athanasiou, and Michael~J Black.
\newblock Vibe: Video inference for human body pose and shape estimation.
\newblock In \emph{Proceedings of the IEEE/CVF conference on computer vision and pattern recognition}, pages 5253--5263, 2020.

\bibitem[Kulon et~al.(2020)Kulon, Guler, Kokkinos, Bronstein, and Zafeiriou]{kulon2020weakly}
Dominik Kulon, Riza~Alp Guler, Iasonas Kokkinos, Michael~M Bronstein, and Stefanos Zafeiriou.
\newblock Weakly-supervised mesh-convolutional hand reconstruction in the wild.
\newblock In \emph{Proceedings of the IEEE/CVF conference on computer vision and pattern recognition}, pages 4990--5000, 2020.

\bibitem[Lee et~al.(2023)Lee, Jang, Kim, Sung, and Kim]{lee2023fourierhandflow}
Jihyun Lee, Junbong Jang, Donghwan Kim, Minhyuk Sung, and Tae-Kyun Kim.
\newblock Fourierhandflow: Neural 4d hand representation using fourier query flow.
\newblock \emph{Advances in Neural Information Processing Systems}, 36:\penalty0 29239--29251, 2023.

\bibitem[Lee et~al.(2024)Lee, Saito, Nam, Sung, and Kim]{lee2024interhandgen}
Jihyun Lee, Shunsuke Saito, Giljoo Nam, Minhyuk Sung, and Tae-Kyun Kim.
\newblock Interhandgen: Two-hand interaction generation via cascaded reverse diffusion.
\newblock In \emph{Proceedings of the IEEE/CVF Conference on Computer Vision and Pattern Recognition}, pages 527--537, 2024.

\bibitem[Li et~al.(2022)Li, An, Zhang, Wu, Chen, Yu, and Liu]{li2022interacting}
Mengcheng Li, Liang An, Hongwen Zhang, Lianpeng Wu, Feng Chen, Tao Yu, and Yebin Liu.
\newblock Interacting attention graph for single image two-hand reconstruction.
\newblock In \emph{Proceedings of the IEEE/CVF Conference on Computer Vision and Pattern Recognition}, pages 2761--2770, 2022.

\bibitem[Li et~al.(2024)Li, Zhang, Zhang, Shao, Yu, and Liu]{li2024hhmr}
Mengcheng Li, Hongwen Zhang, Yuxiang Zhang, Ruizhi Shao, Tao Yu, and Yebin Liu.
\newblock Hhmr: Holistic hand mesh recovery by enhancing the multimodal controllability of graph diffusion models.
\newblock In \emph{Proceedings of the IEEE/CVF Conference on Computer Vision and Pattern Recognition}, pages 645--654, 2024.

\bibitem[Lin et~al.(2021)Lin, Wang, and Liu]{lin2021end}
Kevin Lin, Lijuan Wang, and Zicheng Liu.
\newblock End-to-end human pose and mesh reconstruction with transformers.
\newblock In \emph{Proceedings of the IEEE/CVF Conference on Computer Vision and Pattern Recognition}, pages 1954--1963, 2021.

\bibitem[Liu et~al.(2021)Liu, Jiang, Xu, Liu, and Wang]{liu2021semi}
Shaowei Liu, Hanwen Jiang, Jiarui Xu, Sifei Liu, and Xiaolong Wang.
\newblock Semi-supervised 3d hand-object poses estimation with interactions in time.
\newblock In \emph{Proceedings of the IEEE/CVF Conference on Computer Vision and Pattern Recognition}, pages 14687--14697, 2021.

\bibitem[Liu and Yi(2024)]{liu2024geneoh}
Xueyi Liu and Li Yi.
\newblock Geneoh diffusion: Towards generalizable hand-object interaction denoising via denoising diffusion.
\newblock \emph{arXiv preprint arXiv:2402.14810}, 2024.

\bibitem[Liu et~al.(2024)Liu, Yang, Si, Liu, Li, Zhang, Liu, and Yi]{liu2024taco}
Yun Liu, Haolin Yang, Xu Si, Ling Liu, Zipeng Li, Yuxiang Zhang, Yebin Liu, and Li Yi.
\newblock Taco: Benchmarking generalizable bimanual tool-action-object understanding.
\newblock In \emph{Proceedings of the IEEE/CVF Conference on Computer Vision and Pattern Recognition}, pages 21740--21751, 2024.

\bibitem[Loshchilov(2017)]{loshchilov2017decoupled}
I Loshchilov.
\newblock Decoupled weight decay regularization.
\newblock \emph{arXiv preprint arXiv:1711.05101}, 2017.

\bibitem[Luo et~al.(2024)Luo, Liu, and Yi]{luo2024physics}
Haowen Luo, Yunze Liu, and Li Yi.
\newblock Physics-aware hand-object interaction denoising.
\newblock In \emph{Proceedings of the IEEE/CVF Conference on Computer Vision and Pattern Recognition}, pages 2341--2350, 2024.

\bibitem[Moon and Lee(2020)]{moon2020i2l}
Gyeongsik Moon and Kyoung~Mu Lee.
\newblock I2l-meshnet: Image-to-lixel prediction network for accurate 3d human pose and mesh estimation from a single rgb image.
\newblock In \emph{Computer Vision--ECCV 2020: 16th European Conference, Glasgow, UK, August 23--28, 2020, Proceedings, Part VII 16}, pages 752--768. Springer, 2020.

\bibitem[Mueller et~al.(2018)Mueller, Bernard, Sotnychenko, Mehta, Sridhar, Casas, and Theobalt]{mueller2018ganerated}
Franziska Mueller, Florian Bernard, Oleksandr Sotnychenko, Dushyant Mehta, Srinath Sridhar, Dan Casas, and Christian Theobalt.
\newblock Ganerated hands for real-time 3d hand tracking from monocular rgb.
\newblock In \emph{Proceedings of the IEEE Conference on Computer Vision and Pattern Recognition}, pages 49--59, 2018.

\bibitem[Park et~al.(2022)Park, Oh, Moon, Choi, and Lee]{park2022handoccnet}
JoonKyu Park, Yeonguk Oh, Gyeongsik Moon, Hongsuk Choi, and Kyoung~Mu Lee.
\newblock Handoccnet: Occlusion-robust 3d hand mesh estimation network.
\newblock In \emph{Proceedings of the IEEE/CVF Conference on Computer Vision and Pattern Recognition}, pages 1496--1505, 2022.

\bibitem[Pavlakos et~al.(2023)Pavlakos, Shan, Radosavovic, Kanazawa, Fouhey, and Malik]{pavlakos2023reconstructing}
Georgios Pavlakos, Dandan Shan, Ilija Radosavovic, Angjoo Kanazawa, David Fouhey, and Jitendra Malik.
\newblock Reconstructing hands in 3d with transformers.
\newblock \emph{arXiv preprint arXiv:2312.05251}, 2023.

\bibitem[Potamias et~al.(2024)Potamias, Zhang, Deng, and Zafeiriou]{potamias2024wilor}
Rolandos~Alexandros Potamias, Jinglei Zhang, Jiankang Deng, and Stefanos Zafeiriou.
\newblock Wilor: End-to-end 3d hand localization and reconstruction in-the-wild.
\newblock \emph{arXiv preprint arXiv:2409.12259}, 2024.

\bibitem[Qi et~al.(2024)Qi, Zhao, Salzmann, and Mathis]{Qi_2024_CVPR}
Haozhe Qi, Chen Zhao, Mathieu Salzmann, and Alexander Mathis.
\newblock Hoisdf: Constraining 3d hand-object pose estimation with global signed distance fields.
\newblock In \emph{Proceedings of the IEEE/CVF Conference on Computer Vision and Pattern Recognition (CVPR)}, pages 10392--10402, 2024.

\bibitem[Qin et~al.(2022)Qin, Su, and Wang]{qin2022one}
Yuzhe Qin, Hao Su, and Xiaolong Wang.
\newblock From one hand to multiple hands: Imitation learning for dexterous manipulation from single-camera teleoperation.
\newblock \emph{IEEE Robotics and Automation Letters}, 7\penalty0 (4):\penalty0 10873--10881, 2022.

\bibitem[Ren et~al.(2022)Ren, Zhu, and Zhang]{ren2022end}
Jinwei Ren, Jianke Zhu, and Jialiang Zhang.
\newblock End-to-end weakly-supervised single-stage multiple 3d hand mesh reconstruction from a single rgb image.
\newblock \emph{arXiv preprint arXiv:2204.08154}, 2022.

\bibitem[Ren et~al.(2023)Ren, Wen, Zheng, Xue, Sun, Qi, Wang, and Liao]{ren2023decoupled}
Pengfei Ren, Chao Wen, Xiaozheng Zheng, Zhou Xue, Haifeng Sun, Qi Qi, Jingyu Wang, and Jianxin Liao.
\newblock Decoupled iterative refinement framework for interacting hands reconstruction from a single rgb image.
\newblock In \emph{Proceedings of the IEEE/CVF International Conference on Computer Vision}, pages 8014--8025, 2023.

\bibitem[Romero et~al.(2017)Romero, Tzionas, and Black]{romero2017embodied}
Javier Romero, Dimitrios Tzionas, and Michael~J Black.
\newblock Embodied hands: Modeling and capturing hands and bodies together.
\newblock \emph{ACM Transactions on Graphics (ToG)}, 36\penalty0 (6):\penalty0 1--17, 2017.

\bibitem[Santello et~al.(2013)Santello, Baud-Bovy, and J{\"o}rntell]{santello2013neural}
Marco Santello, Gabriel Baud-Bovy, and Henrik J{\"o}rntell.
\newblock Neural bases of hand synergies.
\newblock \emph{Frontiers in computational neuroscience}, 7:\penalty0 23, 2013.

\bibitem[Song et~al.(2020)Song, Meng, and Ermon]{song2020denoising}
Jiaming Song, Chenlin Meng, and Stefano Ermon.
\newblock Denoising diffusion implicit models.
\newblock \emph{arXiv preprint arXiv:2010.02502}, 2020.

\bibitem[Spurr et~al.(2020)Spurr, Iqbal, Molchanov, Hilliges, and Kautz]{spurr2020weakly}
Adrian Spurr, Umar Iqbal, Pavlo Molchanov, Otmar Hilliges, and Jan Kautz.
\newblock Weakly supervised 3d hand pose estimation via biomechanical constraints.
\newblock \emph{arXiv preprint arXiv:2003.09282}, 2020.

\bibitem[Spurr et~al.(2021)Spurr, Dahiya, Wang, Zhang, and Hilliges]{spurr2021self}
Adrian Spurr, Aneesh Dahiya, Xi Wang, Xucong Zhang, and Otmar Hilliges.
\newblock Self-supervised 3d hand pose estimation from monocular rgb via contrastive learning.
\newblock In \emph{Proceedings of the IEEE/CVF International Conference on Computer Vision}, pages 11230--11239, 2021.

\bibitem[Tevet et~al.(2023)Tevet, Raab, Gordon, Shafir, Cohen-or, and Bermano]{tevet2023human}
Guy Tevet, Sigal Raab, Brian Gordon, Yoni Shafir, Daniel Cohen-or, and Amit~Haim Bermano.
\newblock Human motion diffusion model.
\newblock In \emph{The Eleventh International Conference on Learning Representations}, 2023.

\bibitem[Tripathi et~al.(2023)Tripathi, M{\"u}ller, Huang, Taheri, Black, and Tzionas]{tripathi20233d}
Shashank Tripathi, Lea M{\"u}ller, Chun-Hao~P Huang, Omid Taheri, Michael~J Black, and Dimitrios Tzionas.
\newblock 3d human pose estimation via intuitive physics.
\newblock In \emph{Proceedings of the IEEE/CVF conference on computer vision and pattern recognition}, pages 4713--4725, 2023.

\bibitem[Tse et~al.(2022)Tse, Kim, Leonardis, and Chang]{tse2022collaborative}
Tze Ho~Elden Tse, Kwang~In Kim, Ales Leonardis, and Hyung~Jin Chang.
\newblock Collaborative learning for hand and object reconstruction with attention-guided graph convolution.
\newblock In \emph{Proceedings of the IEEE/CVF Conference on Computer Vision and Pattern Recognition}, pages 1664--1674, 2022.

\bibitem[Tu et~al.(2023)Tu, Huang, Chen, Kang, Bao, Yang, and Yuan]{tu2023consistent}
Zhigang Tu, Zhisheng Huang, Yujin Chen, Di Kang, Linchao Bao, Bisheng Yang, and Junsong Yuan.
\newblock Consistent 3d hand reconstruction in video via self-supervised learning.
\newblock \emph{IEEE Transactions on Pattern Analysis and Machine Intelligence}, 2023.

\bibitem[Vaswani(2017)]{vaswani2017attention}
A Vaswani.
\newblock Attention is all you need.
\newblock \emph{Advances in Neural Information Processing Systems}, 2017.

\bibitem[Vo et~al.(2020)Vo, Sheikh, and Narasimhan]{vo2020spatiotemporal}
Minh Vo, Yaser Sheikh, and Srinivasa~G Narasimhan.
\newblock Spatiotemporal bundle adjustment for dynamic 3d human reconstruction in the wild.
\newblock \emph{IEEE Transactions on Pattern Analysis and Machine Intelligence}, 44\penalty0 (2):\penalty0 1066--1080, 2020.

\bibitem[Wang et~al.(2023)Wang, Zhu, and Wen]{wang2023memahand}
Congyi Wang, Feida Zhu, and Shilei Wen.
\newblock Memahand: Exploiting mesh-mano interaction for single image two-hand reconstruction.
\newblock In \emph{Proceedings of the IEEE/CVF Conference on Computer Vision and Pattern Recognition}, pages 564--573, 2023.

\bibitem[Xie et~al.(2024)Xie, Xu, Tang, Yu, and Lu]{xie2024ms}
Pengfei Xie, Wenqiang Xu, Tutian Tang, Zhenjun Yu, and Cewu Lu.
\newblock Ms-mano: Enabling hand pose tracking with biomechanical constraints.
\newblock In \emph{Proceedings of the IEEE/CVF Conference on Computer Vision and Pattern Recognition}, pages 2382--2392, 2024.

\bibitem[Xu et~al.(2023)Xu, Wang, Tang, and Fu]{xu2023h2onet}
Hao Xu, Tianyu Wang, Xiao Tang, and Chi-Wing Fu.
\newblock H2onet: Hand-occlusion-and-orientation-aware network for real-time 3d hand mesh reconstruction.
\newblock In \emph{Proceedings of the IEEE/CVF Conference on Computer Vision and Pattern Recognition}, pages 17048--17058, 2023.

\bibitem[Xu et~al.(2024)Xu, Li, Wang, Liu, and Fu]{xu2024handbooster}
Hao Xu, Haipeng Li, Yinqiao Wang, Shuaicheng Liu, and Chi-Wing Fu.
\newblock Handbooster: Boosting 3d hand-mesh reconstruction by conditional synthesis and sampling of hand-object interactions.
\newblock In \emph{Proceedings of the IEEE/CVF Conference on Computer Vision and Pattern Recognition}, pages 10159--10169, 2024.

\bibitem[Yang et~al.(2020)Yang, Chang, Lee, and Kwak]{yang2020seqhand}
John Yang, Hyung~Jin Chang, Seungeui Lee, and Nojun Kwak.
\newblock Seqhand: Rgb-sequence-based 3d hand pose and shape estimation.
\newblock In \emph{Computer Vision--ECCV 2020: 16th European Conference, Glasgow, UK, August 23--28, 2020, Proceedings, Part XII 16}, pages 122--139. Springer, 2020.

\bibitem[Yang et~al.(2019)Yang, Li, Lee, and Yao]{yang2019aligning}
Linlin Yang, Shile Li, Dongheui Lee, and Angela Yao.
\newblock Aligning latent spaces for 3d hand pose estimation.
\newblock In \emph{Proceedings of the IEEE/CVF International Conference on Computer Vision}, pages 2335--2343, 2019.

\bibitem[Yang et~al.(2022)Yang, Li, Zhan, Lv, Xu, Li, and Lu]{yang2022artiboost}
Lixin Yang, Kailin Li, Xinyu Zhan, Jun Lv, Wenqiang Xu, Jiefeng Li, and Cewu Lu.
\newblock Artiboost: Boosting articulated 3d hand-object pose estimation via online exploration and synthesis.
\newblock In \emph{Proceedings of the IEEE/CVF Conference on Computer Vision and Pattern Recognition}, pages 2750--2760, 2022.

\bibitem[Ye et~al.(2024)Ye, Gupta, Kitani, and Tulsiani]{Ye_2024_CVPR}
Yufei Ye, Abhinav Gupta, Kris Kitani, and Shubham Tulsiani.
\newblock G-hop: Generative hand-object prior for interaction reconstruction and grasp synthesis.
\newblock In \emph{Proceedings of the IEEE/CVF Conference on Computer Vision and Pattern Recognition (CVPR)}, pages 1911--1920, 2024.

\bibitem[Young and Van~Vliet(1995)]{young1995recursive}
Ian~T Young and Lucas~J Van~Vliet.
\newblock Recursive implementation of the gaussian filter.
\newblock \emph{Signal processing}, 44\penalty0 (2):\penalty0 139--151, 1995.

\bibitem[Yu et~al.(2023{\natexlab{a}})Yu, Huang, Fang, Breckon, and Wang]{yu2023acr}
Zhengdi Yu, Shaoli Huang, Chen Fang, Toby~P Breckon, and Jue Wang.
\newblock Acr: Attention collaboration-based regressor for arbitrary two-hand reconstruction.
\newblock In \emph{Proceedings of the IEEE/CVF Conference on Computer Vision and Pattern Recognition}, pages 12955--12964, 2023{\natexlab{a}}.

\bibitem[Yu et~al.(2023{\natexlab{b}})Yu, Li, Yang, Zheng, Mi, Lee, and Yao]{yu2023overcoming}
Ziwei Yu, Chen Li, Linlin Yang, Xiaoxu Zheng, Michael~Bi Mi, Gim~Hee Lee, and Angela Yao.
\newblock Overcoming the trade-off between accuracy and plausibility in 3d hand shape reconstruction.
\newblock In \emph{Proceedings of the IEEE/CVF Conference on Computer Vision and Pattern Recognition}, pages 544--553, 2023{\natexlab{b}}.

\bibitem[Yue et~al.(2024)Yue, Wang, and Loy]{yue2024resshift}
Zongsheng Yue, Jianyi Wang, and Chen~Change Loy.
\newblock Resshift: Efficient diffusion model for image super-resolution by residual shifting.
\newblock \emph{Advances in Neural Information Processing Systems}, 36, 2024.

\bibitem[Zhang et~al.(2021)Zhang, Wang, Deng, Zhang, Tan, Ma, and Wang]{zhang2021interacting}
Baowen Zhang, Yangang Wang, Xiaoming Deng, Yinda Zhang, Ping Tan, Cuixia Ma, and Hongan Wang.
\newblock Interacting two-hand 3d pose and shape reconstruction from single color image.
\newblock In \emph{Proceedings of the IEEE/CVF International Conference on Computer Vision}, pages 11354--11363, 2021.

\bibitem[Zhang et~al.(2019)Zhang, Li, Mo, Zhang, and Zheng]{zhang2019end}
Xiong Zhang, Qiang Li, Hong Mo, Wenbo Zhang, and Wen Zheng.
\newblock End-to-end hand mesh recovery from a monocular rgb image.
\newblock In \emph{Proceedings of the IEEE/CVF International Conference on Computer Vision}, pages 2354--2364, 2019.

\bibitem[Zhang et~al.(2024{\natexlab{a}})Zhang, Kephart, Cui, and Ji]{zhang2024physpt}
Yufei Zhang, Jeffrey~O Kephart, Zijun Cui, and Qiang Ji.
\newblock Physpt: Physics-aware pretrained transformer for estimating human dynamics from monocular videos.
\newblock In \emph{Proceedings of the IEEE/CVF Conference on Computer Vision and Pattern Recognition}, pages 2305--2317, 2024{\natexlab{a}}.

\bibitem[Zhang et~al.(2024{\natexlab{b}})Zhang, Kephart, and Ji]{zhang2024weakly}
Yufei Zhang, Jeffrey~O Kephart, and Qiang Ji.
\newblock Weakly-supervised 3d hand reconstruction with knowledge prior and uncertainty guidance.
\newblock \emph{arXiv preprint arXiv:2407.12307}, 2024{\natexlab{b}}.

\bibitem[Zhao et~al.(2013)Zhao, Zhang, Min, and Chai]{zhao2013robust}
Wenping Zhao, Jianjie Zhang, Jianyuan Min, and Jinxiang Chai.
\newblock Robust realtime physics-based motion control for human grasping.
\newblock \emph{ACM Transactions on Graphics (TOG)}, 32\penalty0 (6):\penalty0 1--12, 2013.

\bibitem[Zhou et~al.(2022)Zhou, Bhatnagar, Lenssen, and Pons-Moll]{zhou2022toch}
Keyang Zhou, Bharat~Lal Bhatnagar, Jan~Eric Lenssen, and Gerard Pons-Moll.
\newblock Toch: Spatio-temporal object-to-hand correspondence for motion refinement.
\newblock In \emph{European Conference on Computer Vision}, pages 1--19. Springer, 2022.

\bibitem[Zhou et~al.(2024{\natexlab{a}})Zhou, Zhou, Lv, Zou, Tang, and Liang]{Zhou_2024_CVPR}
Zhishan Zhou, Shihao Zhou, Zhi Lv, Minqiang Zou, Yao Tang, and Jiajun Liang.
\newblock A simple baseline for efficient hand mesh reconstruction.
\newblock In \emph{Proceedings of the IEEE/CVF Conference on Computer Vision and Pattern Recognition (CVPR)}, pages 1367--1376, 2024{\natexlab{a}}.

\bibitem[Zhou et~al.(2024{\natexlab{b}})Zhou, Zhou, Lv, Zou, Tang, and Liang]{zhou2024simple}
Zhishan Zhou, Shihao Zhou, Zhi Lv, Minqiang Zou, Yao Tang, and Jiajun Liang.
\newblock A simple baseline for efficient hand mesh reconstruction.
\newblock In \emph{Proceedings of the IEEE/CVF Conference on Computer Vision and Pattern Recognition}, pages 1367--1376, 2024{\natexlab{b}}.

\bibitem[Zhu and Damen(2024)]{zhu2024grip}
Zhifan Zhu and Dima Damen.
\newblock Get a grip: Reconstructing hand-object stable grasps in egocentric videos, 2024.

\bibitem[Ziani et~al.(2022)Ziani, Fan, Kocabas, Christen, and Hilliges]{ziani2022tempclr}
Andrea Ziani, Zicong Fan, Muhammed Kocabas, Sammy Christen, and Otmar Hilliges.
\newblock Tempclr: Reconstructing hands via time-coherent contrastive learning.
\newblock In \emph{2022 International Conference on 3D Vision (3DV)}, pages 627--636. IEEE, 2022.

\bibitem[Zimmermann and Brox(2017)]{zimmermann2017learning}
Christian Zimmermann and Thomas Brox.
\newblock Learning to estimate 3d hand pose from single rgb images.
\newblock In \emph{Proceedings of the IEEE international conference on computer vision}, pages 4903--4911, 2017.

\bibitem[Zuo et~al.(2023)Zuo, Zhao, Sun, Xie, Xue, and Wang]{zuo2023reconstructing}
Binghui Zuo, Zimeng Zhao, Wenqian Sun, Wei Xie, Zhou Xue, and Yangang Wang.
\newblock Reconstructing interacting hands with interaction prior from monocular images.
\newblock In \emph{Proceedings of the IEEE/CVF International Conference on Computer Vision}, pages 9054--9064, 2023.

\end{thebibliography}
}

\clearpage
\setcounter{page}{1}
\setcounter{section}{0}
\maketitlesupplementary
\appendix

In this supplementary material (Appendix as referred to in the main paper), we provide additional evaluation of our proposed approach, including 
\begin{itemize}
    \item Appendix~\ref{appx:fullysupervised}: Evaluation on Model-specific Training
    \item Appendix~\ref{appx:dynamics}: Evaluation on Complex Bimanual Hand-object Manipulation Dataset 
    \item Appendix~\ref{appx:additionalqualitive}: Additional Qualitative Evaluation 
\end{itemize}

\section{Evaluation on Model-specific Training}
\label{appx:fullysupervised}

\begin{table}[h]
\tabcolsep=0.03in
    \begin{center}
    \begin{tabular}{ l cc  ccc }
    \toprule
    \multirow[b]{2}{*}{Method} & \multicolumn{2}{c}{\textit{Rec. Error}} & \multicolumn{3}{c}{\textit{Phys. Plausibility}}  \\ \cmidrule(lr){2-3} \cmidrule{4-6}
    & MJE & P-MJE & ACCL & KIN & STA \\ \cmidrule{1-6}
    HaMer \cite{pavlakos2023reconstructing} & 18.9 & 4.4 & 7.95 & 22.49 & 1.08 \\    
    \cmidrule{1-6}
    VIBE~\cite{kocabas2020vibe} & 17.0 & 6.4 & - & - & - \\
    TCMR~\cite{choi2021beyond} & 16.0 & 6.3 & - & - & - \\
    Deformer~\cite{Fu_2023_ICCV} & 13.6 & 5.2 & - & - & - \\
    BioPR~\cite{choi2021beyond} & 12.9 & - & - & - & - \\
    \cmidrule{1-6}
    \textbf{Ours} (A)  & 17.5 & 4.1 & 1.01 & \textbf{0.00} & \textbf{0.00} \\
    \textbf{Ours} (S) & \textbf{12.4} & \textbf{3.9} & \textbf{0.80} & \textbf{0.00} & \textbf{0.00} \\
    \bottomrule
    \end{tabular}
    \caption{\textbf{Quantitative Comparison of Our Model-Specific Approach against SOTA Methods.} The evaluation is conducted on the DexYCB dataset. Results of other methods are obtained from their published papers. ``\textbf{Ours} (A)" refers to our model-agnostic approach discussed in the main manuscript, while ``\textbf{Ours} (S)" represents our model-specific variant trained to enhance HaMer. For all the values, the smaller, the better.}
    \label{tab:modelspecific}
    \end{center}
\end{table}

In the main paper, we focused on a challenging model-agnostic setting where access to specific frame-wise reconstruction models is not assumed. Here, we explore the potential of our approach through model-specific training, where we adapt our refinement model to enhance a specific frame-wise reconstruction method. Taking the leading approach HaMer as our baseline, we train our model on paired sequences $(\mathbf{x}_{1:T},\mathbf{y}_{1:T})$ consisting of HaMer's frame-wise predictions and ground truth motion data. Table~\ref{tab:modelspecific} shows the evaluation results on the DexYCB dataset. Our model-specific variant (Ours (S)) achieves substantial improvements over HaMer in both reconstruction accuracy (reducing MJE from 18.9 to 12.6 mm) and physical plausibility (reducing ACCL from 7.95 to 0.8 mm/frame$^2$). Moreover, our model-specific variant significantly outperforms recent video-based methods in 3D hand reconstruction accuracy. Specifically, compared to leading method BioPR, our approach achieves better accuracy with an MJE of 12.4 mm versus their 12.9 mm. These improvements further demonstrate the effectiveness of our physics-augmented diffusion-based refinement approach. While our model-agnostic approach provides flexibility across different reconstruction methods, the model-specific training can achieve superior performance when targeting a specific frame-wise reconstruction model.

\section{Evaluation on Complex Bimanual Hand-object Manipulation Dataset}
\label{appx:dynamics}

\begin{table}[h]
\tabcolsep=0.03in
    \begin{center}
    \begin{tabular}{ l c c }
    \toprule
    Method & P-MJE & ACCL \\ \cmidrule{1-3}
    A:HaMer \cite{pavlakos2023reconstructing} & 9.7 & 9.88 \\    
    \cmidrule{1-3}
    A+PoseBERT~\cite{baradel2022posebert} & 11.0 & 3.58 \\
    \cmidrule{1-3}
    A+\textbf{Ours} & \textbf{8.9} & \textbf{0.81} \\
    \bottomrule
    \end{tabular}
    \caption{\textbf{Quantitative Evaluation on Complex Bimanual Hand-object Manipulation Dataset TACO~\cite{liu2024taco}.} For all the values, the smaller, the better.}
    \label{tab:dynamicdata}
    \end{center}
\end{table}

We here extend the evaluation to a recent dataset featuring more complex hand-object interaction sequences to further demonstrate the effectiveness of our approach. Specifically, we consider the TACO dataset, a large-scale benchmark for bimanual hand-object interactions. Captured through a multi-view setup, it encompasses a diverse range of tool-action-object compositions representative of daily human activities. Without any model fine-tuning or retraining, we integrate our method on top of HaMer and evaluate it on the S1 testing split of the TACO dataset. We present the evaluation results in Table~\ref{tab:dynamicdata}. As illustrated, our approach consistently improves upon the leading frame-wise reconstruction HaMer (8.9 vs. 9.7 mm P-MJE, and 0.81 vs. 9.88 mm/frame$^2$ ACCL) and outperforms the leading motion refinement model PoseBERT (11.0 mm P-MJE, and 3.58 mm/frame$^2$ ACCL). These improvements in significant dynamic scenarios involving complex bimanual hand-object manipulation further validate the robustness of our physics-augmented diffusion-based approach.

\section{Additional Qualitative Evaluation}
\label{appx:additionalqualitive}

\begin{figure}[t]
\begin{center}
\includegraphics[width=0.98\linewidth]{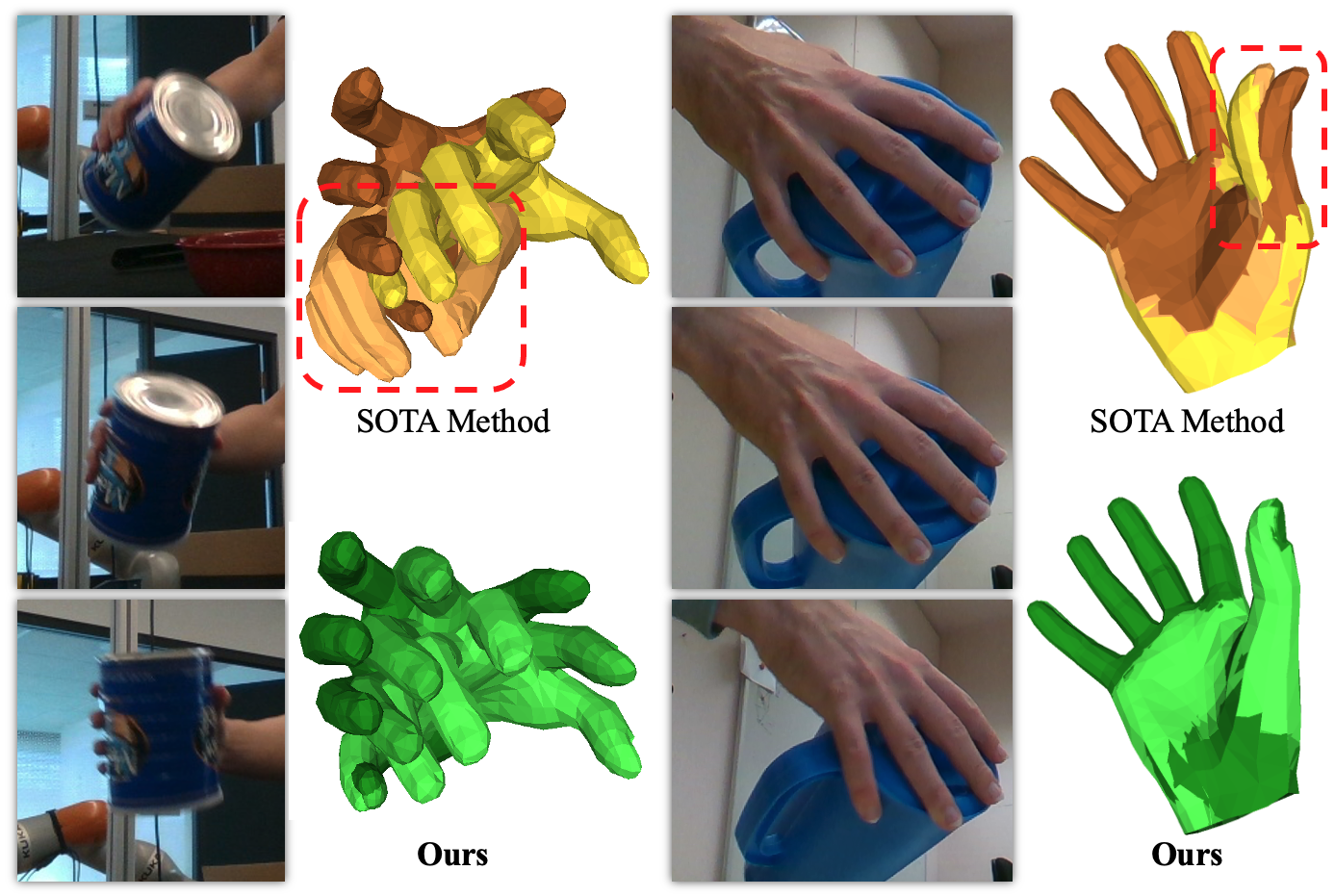}
\caption{\textbf{Qualitative Evaluation on Stable Grasping Sequences.} The testing sequences are from DexYCB (left) and HO3Dv2 (right). To better illustrate the enhanced reconstruction of our model on grasping poses, root rotation is removed in the results on the right.}
\label{fig:ch6-quality}
\end{center}
\end{figure}

Our qualitative evaluation in the main paper demonstrates improved accuracy and physical plausibility in hand motion recovery compared to SOTA approaches. In this section, we further highlight our advantages through evaluation on stable grasping sequences. We present a qualitative comparison with HaMer in Figure~\ref{fig:ch6-quality}. Despite being trained on large-scale data with a high-capacity model, HaMer still produces degraded reconstructions even for simple, static hand poses when partially observed—such as when holding a bottle (highlighted by red rectangles). In contrast, our model achieves temporally consistent motion recovery by capturing an intuitive understanding of how the hand naturally interacts with objects. In particular, the example on the right shows that our method maintains stable hand poses over time during prolonged grasping.

\end{document}